\def\ARXIVVERSION{1}
\documentclass[11pt]{article}

\usepackage[final]{acl}

\usepackage{times}
\usepackage{latexsym}
\usepackage[T1]{fontenc}
\usepackage[utf8]{inputenc}
\usepackage{microtype}
\IfFileExists{inconsolata.sty}{\usepackage{inconsolata}}{} 
\usepackage{multirow}
\usepackage{booktabs}
\usepackage{tabularx}
\usepackage{graphicx}
\usepackage{url}
\usepackage{xcolor}
\usepackage{amssymb}
\usepackage{tikz}
\usetikzlibrary{arrows.meta, positioning, fit, calc, shapes.geometric, backgrounds}
\usepackage{pgfplots}
\pgfplotsset{compat=1.18}

\ifdefined\ARXIVVERSION
  \usepackage{fancyhdr}
  \fancypagestyle{arxivnotice}{%
    \fancyhf{}%
    \fancyfoot[C]{\footnotesize\itshape\shortstack{%
      Accepted at Grounding Language Models: Learning Faithfully and Efficiently (GroundLM 2026),\\
      co-located with EMNLP 2026.}}%
  }
\fi

\hypersetup{
  pdftitle={VERITYGATE: A Four-Gate Schema-Level Faithfulness Framework and Paired Benchmark for Grounded LLM Narrations over Structured Evidence},
  pdfauthor={Sachin Gupta}
}

\newcommand{\veritycoderelease}{\href{https://github.com/sachinkg12/yukti/releases/tag/veritygate-groundlm-2026-v1.0.0}{versioned GitHub code release}}
\newcommand{\verityartifact}{\href{https://doi.org/10.5281/zenodo.22668710}{versioned Zenodo supplementary artifact}}

\definecolor{cbBlue}{RGB}{0,114,178}
\definecolor{cbGreen}{RGB}{0,158,115}
\definecolor{cbOrange}{RGB}{230,159,0}
\definecolor{cbVermillion}{RGB}{213,94,0}
\definecolor{cbPurple}{RGB}{204,121,167}
\definecolor{cbGray}{RGB}{120,120,120}
\definecolor{cbAnnot}{RGB}{90,90,90}

\newcommand{\circnum}[1]{\tikz[baseline=(char.base)]{%
  \node[shape=circle, draw=cbGreen, fill=cbGreen!20, thick,
        inner sep=0.6pt, minimum size=4mm, font=\bfseries\footnotesize]
  (char) {#1};}}

\newcommand{\passmark}{\textcolor{cbGreen}{\textbf{$\checkmark$}}}
\newcommand{\failmark}{\textcolor{cbVermillion}{\textbf{$\times$}}}

\pgfplotsset{
  verity axis/.style={
    axis lines*=left,
    ymajorgrids,
    grid style={dashed, gray!25},
    every axis plot/.append style={fill opacity=0.85, draw opacity=0.95},
    label style={font=\small},
    tick label style={font=\scriptsize},
    legend style={font=\scriptsize, draw=none, fill=none, cells={anchor=west}},
  },
}

\title{\fontsize{13.2}{15.8}\selectfont
VERITYGATE: A Four-Gate Schema-Level Faithfulness Framework and Paired\\
Benchmark for Grounded LLM Narrations over Structured Evidence}

\author{Sachin Gupta \\
  Independent Researcher \\
  San Jose, California, USA \\
  \texttt{sachinkg12@gmail.com}}

\begin{document}
\maketitle
\ifdefined\ARXIVVERSION
  \thispagestyle{arxivnotice}
\fi

\begin{abstract}
Fluent LLM explanations may not follow the evidence from a structured system.
We present VERITYGATE, a four-gate checker for declared evidence IDs, entities,
numbers, and claim types. It checks a fixed schema; it does not verify every
fact in the prose. At $r{=}0$ and $r{=}1$, we test 900 instances per setting
(450 grounded--ungrounded pairs)
with GPT-4o-mini, Llama-3.3-70B, and Claude Sonnet 4.6. Under this schema-level
contract and before repair, 80.3\% of mini claims and 47.9\% of Sonnet claims
fail. These are verifier rejection rates, not prose-hallucination rates. One
repair pass raises claim survival from 19.7\% to 28.0\% for mini and from
52.1\% to 54.3\% for Sonnet. Verified claims per example change by +0.14 for
mini, -0.71 for Llama, and -0.47 for Sonnet, so survival and output volume must
be reported together. A second Sonnet pass gives no clear gain. At r=1, Gate 4
covers 97.0\%, 98.7\%, and 100\% of failing claims for mini, Llama, and Sonnet.
Small human studies support the rules but show gaps between schema
checks and correct prose. A domain-specific GPT-4o judge test shows an order
effect, so it is only a usefulness check. We release the code and
data.
\end{abstract}

\section{Introduction}

The way users interact with structured decision systems is changing. A user
asks for a credit card recommendation, a treatment suggestion, or a loan risk
assessment. A deterministic backend (a solver, a rules engine, a model)
computes the answer. Then a large language model writes the explanation. This
two-stage shape is increasingly common in tool-augmented and retrieval-augmented
production systems for delivering grounded explanations of structured outputs.

The convenience hides a problem. The LLM, even when given the full evidence
graph from the backend, still makes things up. It cites cards that do not
exist. It rounds numbers in ways that change their meaning. It flips signs.
It invents facts that sound plausible.

In production, this is not a research curiosity. In
\textit{Moffatt v.\ Air Canada} (2024 BCCRT 149)~\cite{moffatt2024}, the
airline was held legally liable for false bereavement-fare information its
support chatbot gave to a customer. In March 2024, New York City's MyCity
chatbot (a Microsoft-powered system for small business guidance) was found
advising businesses to take a cut of workers' tips
and lock out tenants, both illegal~\cite{nyc_mycity_2024}. In February 2025, three
attorneys at a top-50 U.S.\ law firm were sanctioned for filing a motion in
which eight of nine cited cases were AI-fabricated~\cite{morgan_morgan_2025}.
As of mid-2026, a public database of court rulings on AI-hallucinated
filings has documented over 1{,}400
decisions~\cite{charlotin_hallucination_db}. The cost of a hallucination in a
regulated sector is no longer hypothetical.

Existing defenses are not enough. Safety alignment within the model is
trained against user-facing harms: jailbreaks, harmful content, direct
prompt injection. It is not trained against ``cite only the data I gave
you.'' Constrained decoding methods like Outlines and JSON schema mode force
syntactic structure but do not check semantic correctness against an external
backend. LLM-as-judge approaches can suffer position effects (our single
judge chooses slot A in 83\% of calls) and, in our diagnostic, prefer
fluent ungrounded narrations over verifier-emitted ones, despite
having no access to the evidence graph.

We present \textsc{VerityGate}, a faithfulness evaluation framework for
grounded LLM narrations, with three contributions:

\begin{enumerate}
    \item \textbf{A claim-level faithfulness taxonomy} for grading
    LLM narrations against a structured backend's evidence: evidence
    existence, entity allowlisting, number binding, and claim-type rule
    satisfaction. The taxonomy and verifier interface are
    designed to be domain-portable and apply to any pipeline where an LLM narrates over
    a structured backend that emits typed evidence with allowed
    entities and numbers; the evidence-graph builder and claim-type
    rules are domain-specific.

    \item \textbf{A paired benchmark} (900 instances at each main setting,
    $r{=}0$ and $r{=}1$: 50 profiles
    $\times$ 3 goals $\times$ 3 narrators $=$ 450 matched settings;
    grounded and ungrounded variants yield 900 instances) with
    pre-registered co-primary metrics and
    paired bootstrap CIs on every contrast. The benchmark enables
    head-to-head comparison of LLM backends and repair strategies under
    a single schema-level structural faithfulness metric
    (\textit{claim-survival rate}).

    \item \textbf{An inference-time repair pipeline} that returns
    failing claims to the LLM with specific verifier errors for a single
    repair pass. No modification to the LLM or the backend. A second
    pass shows no further detectable gain at $r{=}2$ on Sonnet, an
    empirical finding consistent with prior self-correction work.
\end{enumerate}

We additionally contribute a methodology lesson from a counterbalanced
LLM-as-judge protocol with an observed 83\% position-A rate and 0\% tie
rate on one frontier judge: prose-only preference is not a substitute
for mechanical faithfulness verification (Section~\ref{sec:judge}).

We evaluate the pipeline on credit card portfolio optimization as a case
study. The deterministic backend is a mixed-integer linear program (MILP)
that computes the optimal 1--3 card portfolio for a user spending profile;
the LLM narrator writes the explanation; \textsc{VerityGate} sits between them.
We choose MILP as the concrete instantiation because it produces structured
evidence with clean, audit-quality ground truth; the framework requires
only that the backend emit typed evidence with allowed entities and
numbers, a contract we revisit in Section~\ref{sec:discussion}.\footnote{%
The system's name reflects its function: four mechanical gates (one
for each failure mode in the taxonomy) through which every generated
claim must pass before it ships to the user.}

\section{A Four-Gate Verification Taxonomy}
\label{sec:taxonomy}

We catalog the failure modes of LLM narrators when given full structured
grounding. At each of $r{=}0$ and $r{=}1$, 900 narrations form 450 paired
(grounded, ungrounded) comparisons across three LLM backends; four families account for essentially
all observed failures. Each family corresponds to one verification gate.

\paragraph{Gate 1: Evidence Existence.} Every cited evidence
identifier must exist in the structured backend output. LLMs frequently
fabricate identifiers that sound right but do not appear anywhere in
the input. At $r{=}0$ (no repair) Sonnet 4.6 produces 1
evidence-existence error per 150 grounded instances and
\textsc{gpt-4o-mini} produces 6; after one repair pass these drop to 0
and 4 respectively (Table~\ref{tab:main}). The check is deterministic:
a set membership against the backend's emitted identifier set.

\paragraph{Gate 2: Entity Allowlisting.} Cited entity names (card
identifiers, category names, currency codes) must come from the
deployment's allowed entity set. LLMs paraphrase entity names from
training prior, writing ``Amex Blue Cash'' when the backend used
\texttt{amex-bcp}, or inventing ``us-bank-altitude-go'' when no such
card was selected. The check is also deterministic: set membership
against the backend's allowed-entity list.

\paragraph{Gate 3: Number Binding.} Cited number strings (dollar amounts,
percentages, counts) must exactly match a canonical string surfaced from the
structured backend. LLMs round, sign-flip, fabricate, or change decimal
formatting. At $r{=}0$
\textsc{gpt-4o-mini} produces 388 number errors across 150 grounded
instances and Llama-70B produces 182; Sonnet produces 0 at $r{=}0$ and
1 at $r{=}1$. These counts include exact-format mismatches such as
\texttt{2400.0} versus \texttt{2400.00}. Small integers (0--3) are allowed
by convention. The evaluated check is exact string membership.

\paragraph{Gate 4: Claim-Type Rules.} Claims of a given type must
cite evidence of the required type. A comparison claim must cite a
winner-by-category evidence node; a fee-justification claim must cite a
fee-break-even node; a cap-switch claim must cite cap-hit evidence and,
when those node types are present in the graph, both cap-hit and
allocation-segment evidence. LLMs frequently produce structurally
well-formed claims that cite the wrong evidence category. This is the
hardest gate in our evaluation: on Sonnet 4.6 the count drops only from
985 ($r{=}0$) to 841 ($r{=}1$) across 150 instances, and a second repair
pass moves it no further (Section~\ref{sec:eval}). Across these three
models and repair settings, this persistence suggests a type-routing
bottleneck; broader model and domain studies are needed.

\begin{figure}[t]
\centering
\begin{tikzpicture}[
  font=\small,
  trace box/.style={
    rectangle, rounded corners=4pt, draw, thick,
    inner sep=8pt, font=\small,
    text width=7cm,
    align=left,
  },
  passbox/.style={trace box, draw=cbGreen, fill=cbGreen!5},
  failbox/.style={trace box, draw=cbVermillion, fill=cbVermillion!5},
]

\node[passbox] (passbox) {
  \textbf{\textcolor{cbGreen}{Passing claim}}\\[2pt]
  ``\emph{us-bank-altitude-go earns \$1031.00, credits \$0.00, fee \$0.00, net \$1031.00}''\\[5pt]
  \textbf{Type:} \texttt{FEE\_JUSTIFICATION}\\
  \textbf{Cited entities:} \texttt{\{us-bank-altitude-go\}}\\
  \textbf{Cited numbers:} \texttt{\{1031.00\}}\\[4pt]
  \begin{tabular}{@{}p{5.4cm}l@{}}
    {\scriptsize G1 cited evidence \texttt{ev\_5} in graph} & \passmark\\
    {\scriptsize G2 \texttt{us-bank-altitude-go} in allowed entities} & \passmark\\
    {\scriptsize G3 \texttt{1031.00} in allowed numbers} & \passmark\\
    {\scriptsize G4 \texttt{ev\_5} has type \texttt{FEE\_BREAK\_EVEN}} & \passmark\\
  \end{tabular}\\[4pt]
  \textbf{\textcolor{cbGreen}{$\Rightarrow$ VERIFIED: ships to user}}
};

\node[failbox, below=6pt of passbox] (failbox) {
  \textbf{\textcolor{cbVermillion}{Rejected claim}}\\[2pt]
  ``\emph{citi-strata-elite wins DINING (delta \$691.60 over amex-gold)}''\\[5pt]
  \textbf{Type:} \texttt{COMPARISON}\\
  \textbf{Cited entities:} \texttt{\{citi-strata-elite, DINING, amex-gold\}}\\
  \textbf{Cited numbers:} \texttt{\{691.60\}}\\[4pt]
  \begin{tabular}{@{}p{5.4cm}l@{}}
    {\scriptsize G1 cited evidence \texttt{ev\_7} in graph} & \passmark\\
    {\scriptsize G2 \texttt{amex-gold} \textbf{not} in allowed entities} & \failmark\\
    {\scriptsize G3 \texttt{691.60} in allowed numbers} & \passmark\\
    {\scriptsize G4 \texttt{COMPARISON} cites \texttt{WINNER\_BY\_CATEGORY}} & \passmark\\
  \end{tabular}\\[4pt]
  \textbf{\textcolor{cbVermillion}{$\Rightarrow$ REJECTED at Gate~2:}}\\
  {\scriptsize\ttfamily\ \ ``citedEntity not allowed: amex-gold''}\\[3pt]
  {\scriptsize\textcolor{cbAnnot}{\textit{$\rightarrow$ returned to LLM with this specific error for a repair pass.}}}
};

\end{tikzpicture}
\caption{Illustrative traces reconstructed from claim text observed in our
paper-grade evaluation; identifiers and structured declarations are simplified
to isolate one gate at a time. \textit{Top:} a \texttt{FEE\_JUSTIFICATION} claim
passes all four gates and ships. \textit{Bottom:} a \texttt{COMPARISON}
claim fails Gate~2 (the cited entity \texttt{amex-gold} is not in the
allowed-entity list), is rejected with an explicit error message, and
    is returned to the LLM for a single repair pass. These are pedagogical
    traces rather than exact serialized log records. This is the
verification contract that mechanically enforces structural
faithfulness at the schema level.}
\label{fig:trace}
\end{figure}
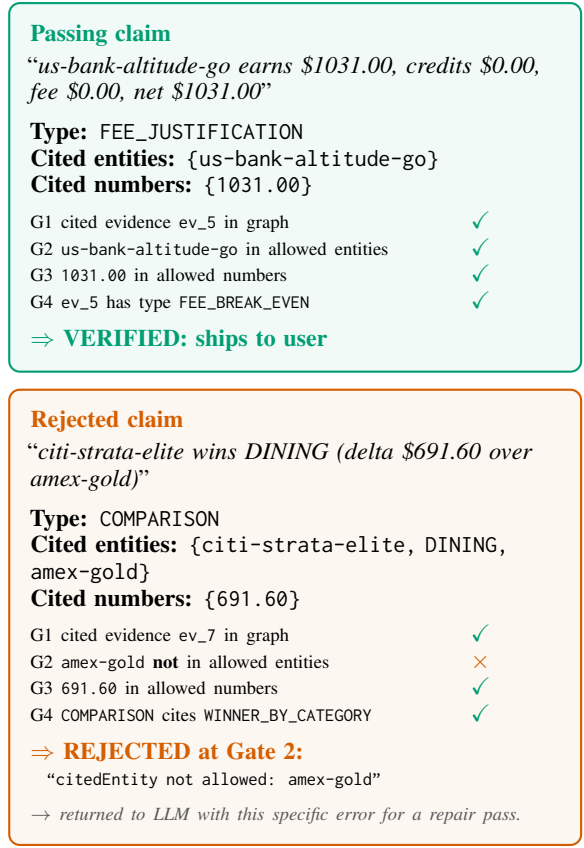

Figure~\ref{fig:trace} shows two illustrative traces reconstructed from
claim text observed in our paper-grade evaluation, making
the four-gate contract concrete before we describe the framework
architecture (Section~\ref{sec:framework}).

Table~\ref{tab:taxonomy} summarizes the four gates and their empirical
profiles.

\begin{table*}[t]
\centering
\small
\setlength{\tabcolsep}{4pt}
\begin{tabularx}{\textwidth}{@{}p{3.25cm}Xp{3.25cm}p{4.25cm}@{}}
\toprule
\textbf{Gate} & \textbf{What it checks} & \textbf{Detection} & \textbf{Empirical grounded error range (150 inst. per cell)} \\
\midrule
1. Evidence Existence & Cited evidence id is in the graph & Set membership & $r{=}0$: 1--6 \quad $r{=}1$: 0--4 (Sonnet best, mini worst) \\
2. Entity Allowlisting & Cited entity is in allowed set & Set membership & $r{=}0$: 0--47 \quad $r{=}1$: 0--43 (Sonnet best, mini worst) \\
3. Number Binding & Cited number string is in allowed set & Exact string membership & $r{=}0$: 0--388 \quad $r{=}1$: 1--97 (Sonnet best, mini worst) \\
4. Claim-Type Rules & Claim type cites required evidence type & Type-rule check & $r{=}0$: 885--985 \quad $r{=}1$: 652--841 (all tiers high) \\
\bottomrule
\end{tabularx}
\caption{The four verification gates. Gates 1--3 are largely addressed by
allowlist surfacing plus one repair pass on the frontier model. Gate~4 is
the persistent residual mode in the evaluated cells: every tier produces
hundreds of type-rule errors after repair, and a second Sonnet repair pass
does not significantly reduce them (Section~\ref{sec:eval}).}
\label{tab:taxonomy}
\end{table*}

\section{The \textsc{VerityGate} Framework}
\label{sec:framework}

\textsc{VerityGate} operates as a transparent layer between the deterministic
backend and the LLM narrator. It requires no modification to either side. The
backend emits structured evidence in its native format; the LLM is called
with a standard prompt. The framework performs three-stage validation.

\subsection{Stage 1: Evidence Graph Construction}

We build an evidence graph from the backend output. Each evidence block
becomes a node; edges connect logically related nodes (for example, a
cap-hit node depends on the winner-by-category node for the same category).
The graph also carries two allowlists:

\begin{itemize}
    \item \textbf{Allowed entities:} all entity strings that appear in the
    backend output (card identifiers, category names, currency codes).
    \item \textbf{Allowed numbers:} all numbers that appear in any evidence
    node's structured fields, plus a small-integer allowlist (0, 1, 2, 3).
\end{itemize}

The graph carries a SHA-256 digest of its canonical form. This makes the
verification step deterministic and auditable: any two runs over the same
backend output produce the same digest.

\subsection{Stage 2: Four-Gate Verification}

The narrator LLM is prompted (strict JSON) to emit prose plus a
parallel list of structured claim objects with declared
\texttt{citedEvidenceIds}, \texttt{citedEntities}, and
\texttt{citedNumbers} (schema and Gate~4 type rules in
Appendix~\ref{app:schema}; prompt templates in
Appendix~\ref{app:prompts}). The four gates check these declared
structured fields against the evidence graph; the \textbf{guarantee}
is \emph{schema-level structural correctness}, not prose-level
faithfulness: an under-citing LLM can mention unverified entities or
numbers in prose that the verifier never inspects
(Appendix~\ref{app:rater-validation}).

Each claim is checked against the four gates from
Section~\ref{sec:taxonomy}. The check is purely structural: no LLM is
involved. If any claim fails any gate, the claim is rejected.
Crucially, rejection is independent of fluency. A grammatically beautiful,
plausible-sounding claim that cites a non-existent number is rejected just
like one containing an unallowed declared value.

The output of Stage 2 is two lists: \textit{verified claims} (passed all four
gates) and \textit{failing claims} (with attached error messages indicating
which gate(s) failed and which strings were unallowed).

\subsection{Stage 3: Repair Loop and Permissive Emission}

If any claim failed verification, the framework can optionally re-prompt
the LLM with the original context plus the specific verifier errors
(``Your previous claims failed for these reasons. Use only the allowed
strings.''). The model produces a new claim list, which is verified
again. We keep whichever of the initial and post-repair lists has fewer
failing claims (intra-pass selection; distinct from the deployment-time
best-of-($r{=}0, r{=}1$) policy in Appendix~\ref{app:prereg}).

We use one repair pass as the headline operating point. On Sonnet, a second
pass produces a paired $\Delta$ that is not statistically distinguishable
from zero (Section~\ref{sec:eval}). For Llama, one pass reduces shipped
volume without a detectable survival gain, so deployments should select the
operating point by model and report both co-primary metrics.

The final output is the \textit{verified claim subset}: only claims
that pass all four gates ship to the user, as their concatenated
\texttt{text} fields. We call this \textit{permissive emission}: emit
whatever passes, drop what fails. The user never sees a failing claim.
Permissive emission is a \emph{filtering} policy, not a
generation-time correction: the underlying LLM continues to fabricate;
claim-survival measures what passes verification (what the user sees),
not generator improvement.

\begin{figure*}[t]
\centering
\begin{tikzpicture}[
  >={Latex[length=2.5mm, width=1.8mm]},
  font=\small,
  node distance=0.6cm and 1.8cm,
  box/.style={
    rectangle, rounded corners=3pt, draw, thick,
    minimum height=0.85cm, align=center, inner sep=5pt, font=\small
  },
  backend/.style={box, fill=cbBlue!15, draw=cbBlue, minimum width=2.6cm},
  llm/.style={box, fill=cbOrange!20, draw=cbOrange, minimum width=2.6cm, minimum height=1.4cm},
  user/.style={box, fill=cbGray!15, draw=cbGray, minimum width=4.4cm},
  stage/.style={
    rectangle, rounded corners=3pt, draw=cbGreen, thick,
    fill=white, minimum width=5.6cm, minimum height=0.95cm,
    align=center, inner sep=5pt, font=\small
  },
  stagewide/.style={stage, minimum height=1.7cm, fill=white},
  gate/.style={
    rectangle, rounded corners=2pt, draw=cbGreen, fill=cbGreen!22,
    minimum width=1.05cm, minimum height=0.7cm,
    font=\scriptsize, align=center, inner sep=2pt
  },
  arrlabel/.style={font=\scriptsize, fill=white, inner sep=1pt},
  annot/.style={
    font=\scriptsize\itshape, text=cbAnnot, align=left, anchor=west,
    text width=3.2cm
  },
]

\node[backend, align=center] (backend) {Deterministic Backend\\\scriptsize{(MILP, rules engine, or model)}};

\node[stage, below=0.8cm of backend] (s1) {\circnum{1}\; Build Evidence Graph};

\node[stagewide, below=0.7cm of s1] (s2) {};
\node[anchor=north, yshift=-3pt] at (s2.north) (s2title) {\circnum{2}\; Four-Gate Verification};
\node[gate, below=4pt of s2title, xshift=-1.85cm] (g1) {\textbf{G1}\\evidence};
\node[gate, right=3pt of g1] (g2) {\textbf{G2}\\entity};
\node[gate, right=3pt of g2] (g3) {\textbf{G3}\\number};
\node[gate, right=3pt of g3] (g4) {\textbf{G4}\\type};

\node[stage, below=0.7cm of s2] (s3) {\circnum{3}\; Permissive Emission};

\node[llm, right=2.4cm of s2] (llm) {LLM\\Narrator\\\scriptsize{(probabilistic)}};

\node[user, below=0.8cm of s3] (user) {Verified Claims to User};

\node[annot, right=0.5cm of s1] (a1)
  {typed evidence nodes,\\allowed entities + numbers,\\SHA-256 canonical digest};
\node[annot, right=0.5cm of s3] (a3)
  {drop failing claims,\\ship only the verified set};

\begin{scope}[on background layer]
  \node[
    draw=cbGreen, thick, dashed, rounded corners=6pt,
    fit=(s1)(s2)(s3),
    fill=cbGreen!6,
    inner xsep=14pt, inner ysep=14pt,
    label={[anchor=south east, font=\bfseries\small, text=cbGreen, xshift=-4pt, yshift=2pt]
           north west:\textsc{VerityGate}}
  ] (vg) {};
\end{scope}

\draw[->, thick] (backend) -- (s1)
  node[arrlabel, midway, right] {structured evidence};
\draw[->, thick] (s1) -- (s2)
  node[arrlabel, midway, right] {evidence graph};
\draw[->, thick] (s2) -- (s3)
  node[arrlabel, midway, right] {verified subset};
\draw[->, thick] (s3) -- (user)
  node[arrlabel, midway, right] {verified claims};

\draw[->, thick]
  ($(s2.east) + (0, 0.35)$) -- ($(llm.west) + (0, 0.35)$)
  node[arrlabel, midway, above] {prompt + graph};
\draw[->, thick]
  ($(llm.west) + (0, 0)$) -- ($(s2.east) + (0, 0)$)
  node[arrlabel, midway, above] {claim JSON};
\draw[->, very thick, dash pattern=on 3pt off 2pt, color=cbVermillion]
  ($(s2.east) + (0, -0.35)$) -- ($(llm.west) + (0, -0.35)$)
  node[arrlabel, midway, below, text=cbVermillion, font=\scriptsize\bfseries] {failing + errors};

\end{tikzpicture}
\caption{Three-stage interception architecture of \textsc{VerityGate},
operating as a transparent layer between the deterministic backend (here a
MILP optimizer) and the LLM narrator. Stage~1 constructs the evidence graph
with typed nodes, an allowed-entity set, and an allowed-number set. Stage~2
verifies each parsed claim against four gates (G1: evidence existence,
G2: entity allowlist, G3: number binding, G4: claim-type rules); the
optional repair loop (vermillion dashed arrow) re-prompts the LLM with the
verifier errors for one pass. Stage~3 is a pure filter: only claims
passing all four gates ship.}
\label{fig:arch}
\end{figure*}
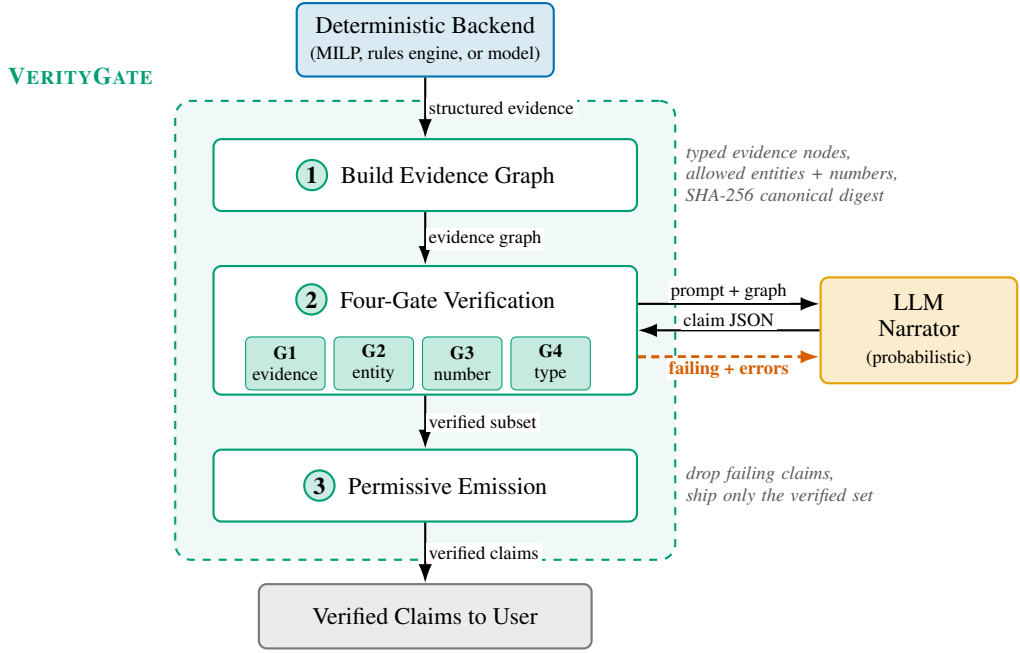

Figure~\ref{fig:arch} shows the overall flow.

\section{Evaluation}
\label{sec:eval}

We evaluate \textsc{VerityGate} along three dimensions: claim-survival rate (the
share of parsed-and-retained claims that pass verification), shipped verified claims per
instance (the user-visible output), and judge-based usefulness preference (a
secondary metric).

\subsection{Setup}

\paragraph{Case study and profiles.}
We test on credit card portfolio optimization. The deterministic backend
is a MILP that emits six task-facing evidence types (cap-hit,
winner-by-category, allocation-segment, fee-break-even, earn-rate,
assumption); the schema also permits four support/container types. It is
independently validated on 1{,}000 real households drawn from the U.S.\
BLS Public-Use Microdata~\citep{bls_cex_pumd} (Spearman $\rho = 0.896$ vs ground-truth optimal
portfolios across 25 survey waves). We rely on that validation for the
credibility of the structured ground truth and do not re-derive it here.
For this evaluation we use 50 parametric profiles $\times$ 3 goal types
(cashback, flexible points, program points) $=$ 150 (profile, goal)
instances, fixed seed 42, stratified to include cap-boundary cases.
Each of $r{=}0$ and $r{=}1$ has $150 \times 3$ narrators
$\times 2$ variants $=$ \textbf{900 benchmark instances}, forming 450 matched
grounded--ungrounded pairs. We add a Sonnet-only ablation at $r{=}2$
(300 more). The released result records contain 16{,}180 parsed-and-retained claim
objects across these conditions. The verifier-passing subset is
what we report as ``shipped'' in Table~\ref{tab:main}.

\paragraph{LLM backends.} Three narrators spanning the capability range:
\textsc{gpt-4o-mini-2024-07-18} (OpenAI, small closed),
\textsc{Llama-3.3-70B-Instruct-Turbo} (Meta, via Together AI,
open-weights mid), and \textsc{Claude Sonnet 4.6} (Anthropic, frontier
closed). The judge model is \textsc{gpt-4o-2024-08-06}: a different
model from the smallest narrator (\textsc{gpt-4o-mini}) within the same
vendor, and a different family entirely from the frontier
narrator.\footnote{All API calls used temperature 0.0 with no
provider-side caching; calls were made between 2026-03-01 and 2026-05-14.
Cost estimates (Section~\ref{sec:cost}) use vendor list pricing as of
2026-05-14, with per-call token counts measured from API responses.}

\paragraph{Variants.} For each (profile, goal, model) we run two paired
variants. \textit{Grounded}: prompt includes the full structured evidence,
the allowed entity list, and the allowed number list. \textit{Ungrounded}:
prompt includes only the goal name, the portfolio card list, and the
top-line breakdown (matching the engine's production
\texttt{UngroundedLlmNarrator}).

\paragraph{Two runs, plus an ablation.} We perform two paired runs:
$r{=}0$ (no repair, baseline) and $r{=}1$ (one repair pass, our headline
configuration). We also run a Sonnet-only ablation at $r{=}2$ to confirm
diminishing returns.

\paragraph{Pre-registration.}
Co-primary metrics, pairing units, statistics, and explicit non-claims
were pre-registered; see Appendix~\ref{app:prereg}. The protocol called the
survival denominator ``generated claims.'' The evaluation used a lenient
parser that skipped malformed individual records and marked an instance
\texttt{schemaFailed} only if no record remained. Thus \texttt{totalClaims}
counts parsed-and-retained claims; a fully failed instance ships zero but adds
zero claims to this denominator. Pre-parse arrays were not archived, so the
partial-record drop rate cannot be recovered.

\paragraph{Statistical procedure.} All confidence intervals use cluster
bootstrap on instances (the natural unit, since claims within an instance
share a profile/goal/model), 10{,}000 iterations, seed 42. We primarily
report paired differences. For repair-effect analysis, pairs are matched
by (profile, goal, model). For cross-model contrasts, pairs are matched
by (profile, goal).

\paragraph{Baselines.} We compare three operating points: (i)
\textit{verify+filter only} ($r{=}0$ in Table~\ref{tab:main}); (ii)
\textit{full \textsc{VerityGate}} ($r{=}1$: verify+filter+repair);
(iii) \textit{ungrounded narration} (high-level summary only, every
``Ungrounded'' row). A \textit{no-defense} (ship-all, no filter)
baseline is not run; the complement of the $r{=}0$ survival fraction
quantifies the user-visible failure rate (e.g., 80.3\% at
\textsc{mini} grounded $r{=}0$). We \emph{do not} evaluate LLM-as-judge
as a \emph{filter} baseline: our judge study
(Section~\ref{sec:judge}) finds that GPT-4o prefers ungrounded narrations
over verified ones and chooses slot A in 83\% of calls. Because it sees
no evidence graph, we do not treat it as a claim-level filter. Instead,
the judge is a diagnostic on
\emph{whether automatic preference signals track faithfulness}: they
do not.

\subsection{Main results}

Table~\ref{tab:main} reports per-cell claim-survival rate, shipped verified
claims per instance, and four-gate breakdown at both $r{=}0$ and $r{=}1$.
Across all three model tiers, the grounded variant under repair ships
more verified claims than the ungrounded variant; the gap widens with
model capability. On the frontier model, the grounded variant under
repair ships 6.67 verified claims per instance.

\begin{table*}[t]
\centering
\small
\setlength{\tabcolsep}{4pt}
\begin{tabular}{llrrrrrrrr}
\toprule
 & & \multicolumn{2}{c}{\textbf{Claim survival}} & \multicolumn{2}{c}{\textbf{Shipped / inst.}} & \multicolumn{4}{c}{\textbf{Failing claims per gate ($r{=}1$, total)}} \\
\cmidrule(lr){3-4} \cmidrule(lr){5-6} \cmidrule(lr){7-10}
\textbf{Model} & \textbf{Variant} & $r{=}0$ & $r{=}1$ & $r{=}0$ & $r{=}1$ & Evid. & Ent. & Num. & Type \\
\midrule
\multirow{2}{*}{\textsc{gpt-4o-mini}} & Grounded & 19.7\% & \textbf{28.0\%} & 1.60 & 1.74 & 4 & 43 & 97 & 652 \\
                                     & Ungrounded & 26.1\% & 25.6\% & 1.29 & 1.31 & 4 & 27 & 263 & 486 \\
\multirow{2}{*}{\textsc{Llama-3.3-70B}} & Grounded & 40.1\% & 41.3\% & 4.29 & 3.58 & 0 & 3 & 84 & 753 \\
                                     & Ungrounded & 21.6\% & 22.0\% & 0.99 & 1.02 & 0 & 344 & 334 & 227 \\
\multirow{2}{*}{\textsc{Claude Sonnet 4.6}} & Grounded & 52.1\% & \textbf{54.3\%} & 7.15 & 6.67 & 0 & 0 & 1 & 841 \\
                                     & Ungrounded & 43.2\% & 41.6\% & 2.36 & 2.21 & 0 & 46 & 196 & 294 \\
\bottomrule
\end{tabular}
\caption{Main results for 900 benchmark instances at each repair setting
(50 profiles $\times$ 3 goals $\times$ 3 narrators $\times$ 2 variants).
Results are separate for $r{=}0$ and $r{=}1$. Conditions:
\textit{Grounded}/$r{=}0$ = verify+filter, no repair pass;
\textit{Grounded}/$r{=}1$ = full \textsc{VerityGate}
(verify+filter+repair); \textit{Ungrounded}/$r$ = same pipeline,
ungrounded prompt. The $r{=}0$ and $r{=}1$ columns are per-pass raw
aggregates; the deployment policy (Appendix~\ref{app:prereg}) is
per-instance best-of selection. Claim survival = passing claims /
parsed-and-retained claims. ``Shipped / inst.'' is mean verified claims
per instance. The last four columns report \textbf{distinct failing claims
per gate} at $r{=}1$ (once within each gate). A claim may appear in
multiple gate columns, so their sum is not a distinct-claim total.
Table~\ref{tab:gate4-submode} separately breaks Gate~4 into
\emph{verifier messages} by claim type.}
\label{tab:main}
\end{table*}

\begin{figure}[t]
\centering
\begin{tikzpicture}
\begin{axis}[
  verity axis,
  ybar,
  bar width=10pt,
  width=\linewidth,
  height=5cm,
  ymin=0, ymax=80,
  ylabel={Claim survival rate (\%)},
  tick label style={font=\small},
  symbolic x coords={mini, Llama-70B, Sonnet},
  xtick=data,
  enlarge x limits=0.4,
  nodes near coords={\pgfmathprintnumber[fixed, precision=0]{\pgfplotspointmeta}},
  nodes near coords style={font=\scriptsize, yshift=8pt, anchor=south, /pgf/number format/fixed},
  error bars/y dir=both,
  error bars/y explicit,
  error bars/error bar style={thick, black!70},
  legend style={
    font=\footnotesize, draw=black!40, fill=white, fill opacity=0.95,
    text opacity=1, cells={anchor=west}, at={(0.02,0.98)}, anchor=north west
  },
  legend cell align=left,
]
\addplot[draw=cbBlue, fill=cbBlue!25]
  coordinates {
    (mini,       19.7) +- (0, 2.7)
    (Llama-70B,  40.1) +- (0, 2.2)
    (Sonnet,     52.1) +- (0, 1.5)
  };
\addplot[draw=cbGreen, fill=cbGreen!30]
  coordinates {
    (mini,       28.0) +- (0, 3.5)
    (Llama-70B,  41.3) +- (0, 2.6)
    (Sonnet,     54.3) +- (0, 2.0)
  };
\legend{$r{=}0$ (no repair), $r{=}1$ (\textsc{VerityGate})}
\end{axis}
\end{tikzpicture}
\caption{Claim-survival rate by model and repair setting, with 95\%
cluster-bootstrap CIs (n=150 per cell). Repair helps the smallest model
most; Llama's change is inconclusive; Sonnet gains marginally.}
\label{fig:survival}
\end{figure}
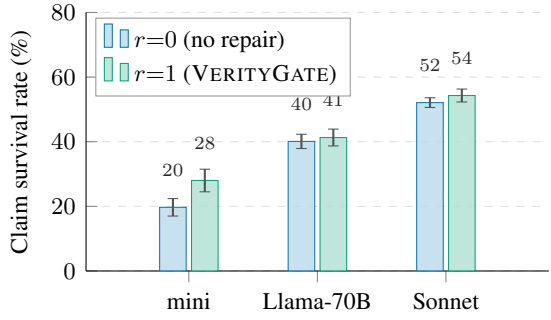

Figure~\ref{fig:survival} visualizes the per-model survival rates across
both repair settings, with bootstrap CI bars.

\paragraph{Per-model repair effect (paired, $r{=}1 - r{=}0$).}
Table~\ref{tab:repair} shows the paired bootstrap differences. Repair
helps small models a lot, frontier models a little, and the open-weights
mid-tier not significantly.

\begin{table}[t]
\centering
\small
\begin{tabular}{lrr}
\toprule
\textbf{Model} & $\Delta$ \textbf{survival} & \textbf{Sig.} \\
\midrule
\textsc{gpt-4o-mini} & \textbf{+8.3 pts} [+4.3, +12.5] & yes \\
\textsc{Llama-3.3-70B} & +1.2 pts [-1.3, +3.6] & no \\
\textsc{Claude Sonnet 4.6} & +2.2 pts [+0.4, +4.1] & yes \\
\bottomrule
\end{tabular}
\caption{Paired $\Delta$ ($r{=}1 - r{=}0$) on grounded claim survival,
95\% cluster bootstrap CI, paired by (profile, goal, model). Llama shows
no detectable survival gain under one pass. Paired $\Delta$ on
shipped verified claims per instance: mini $+0.14$ [$-0.17, +0.47$] (NS);
Llama $-0.71$ [$-1.05, -0.37$] (down significantly, consistent with
becoming more conservative under repair); Sonnet $-0.47$ [$-0.86, -0.08$]
(down significantly).}
\label{tab:repair}
\end{table}

\paragraph{Cross-model contrasts at $r{=}1$.} The three-tier ordering
is clean with non-overlapping CIs: Sonnet $-$ mini $=$ +26.4 pts
[+22.6, +30.1], Llama $-$ mini $=$ +13.3 pts [+9.2, +17.4], and Sonnet
$-$ Llama $=$ +13.0 pts [+10.3, +15.7].

\paragraph{Grounded vs ungrounded: rate is not volume.} On mini at
$r{=}0$, the ungrounded variant has a \emph{higher} per-claim survival
rate (26.1\%) than the grounded variant (19.7\%) even though it ships
\emph{fewer} verified claims per instance (1.29 vs 1.60). Without an
evidence graph the model produces a shorter, safer narration that
references fewer entities and numbers, so each individual claim is less
likely to fail any gate, but the user receives less verifiable
content overall. The deployment-relevant metric is therefore shipped
verified claims per instance, not survival rate alone, which is why we
pre-registered both as co-primary.

\paragraph{Diminishing returns and honesty under pressure.}
A second repair pass on Sonnet gives
$\Delta$($r{=}2 - r{=}1$) $=$ +0.77 pts [$-1.22, +2.81$] (CI crosses
zero), so we deploy at $r{=}1$. Capable models become more conservative
under repair: mean shipped count drops slightly on Llama ($\Delta =
-0.71$) and Sonnet ($\Delta = -0.47$), abandoning claims they cannot
substantiate rather than producing wrong ones. At $r{=}1$, Sonnet still
ships 6.67 verified claims per instance, more than either other evaluated
model; its own $r{=}0$ cell ships 7.15.

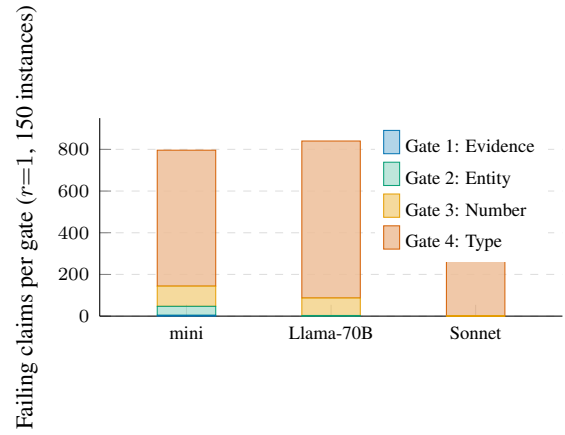
\begin{figure}[t]
\centering
\begin{tikzpicture}
\begin{axis}[
  verity axis,
  ybar stacked,
  bar width=22pt,
  width=\linewidth,
  height=4.2cm,
  ylabel={Failing claims per gate ($r{=}1$, 150 instances)},
  symbolic x coords={mini, Llama-70B, Sonnet},
  xtick=data,
  enlarge x limits=0.3,
  legend pos=north east,
  legend style={font=\scriptsize, fill=white, draw=none},
  legend cell align=left,
  ymin=0, ymax=950,
  ytick={0,200,400,600,800},
]
\addplot[fill=cbBlue!35, draw=cbBlue]
  coordinates {(mini,4) (Llama-70B,0) (Sonnet,0)};
\addplot[fill=cbGreen!30, draw=cbGreen]
  coordinates {(mini,43) (Llama-70B,3) (Sonnet,0)};
\addplot[fill=cbOrange!45, draw=cbOrange]
  coordinates {(mini,97) (Llama-70B,84) (Sonnet,1)};
\addplot[fill=cbVermillion!40, draw=cbVermillion]
  coordinates {(mini,652) (Llama-70B,753) (Sonnet,841)};
\legend{Gate 1: Evidence, Gate 2: Entity, Gate 3: Number, Gate 4: Type}
\end{axis}
\end{tikzpicture}
\caption{Failing claims per gate on the grounded variant at $r{=}1$
(150 instances per model; claims are counted once within each gate and
may appear under multiple gates). Gate~4 (claim-type rules) dominates on
every tier and is the residual problem after repair nearly eliminates
Gates 1--3 on the frontier model. On Sonnet, all 841 distinct failing
claims violate Gate~4; one also violates Gate~3. This is a persistent
residual mode in the tested cells, not a universal scaling claim.}
\label{fig:gate-breakdown}
\end{figure}

\paragraph{Where the failures live.}
Figure~\ref{fig:gate-breakdown} visualizes the gate-error breakdown on
the grounded variant under one repair pass. Gate~4 (claim-type rules)
dominates across all three tiers: it affects 652/672 (97.0\%), 753/763
(98.7\%), and 841/841 (100\%) distinct failing claims on mini, Llama,
and Sonnet, respectively. The sole Sonnet Gate~3 failure overlaps a
Gate~4 failure. Gates 1--3 are essentially eliminated on Sonnet
under repair (0 evidence errors, 0 entity errors, 1 number error across
150 instances). Within the tested models, the residual bottleneck is
type categorization rather than allowlisting or number-grounding.

\paragraph{Structural-only verification: prose-vs-citation gap.}
A small external rater study (Appendix~\ref{app:rater-validation})
surfaces two limitations distinct from Gate~4. On four items, all three
raters accepted the prose values while the verifier rejected the declared
number strings, mainly because numerically equivalent forms such as
\texttt{2400.0} and \texttt{2400.00} did not match exactly; one item also
failed a Gate~4 rule. In the opposite direction, one unanimous
\textsc{No} vs verifier \textsc{Yes} case contained an undeclared prose
phrase and empty cited fields, leaving the verifier nothing to inspect.
Together, these cases motivate combined \textit{structural + prose}
verification and explicit numeric canonicalization as future work.

\subsection{Counterbalanced Judge Analysis}
\label{sec:judge}

We additionally evaluate user-facing usefulness via a counterbalanced
LLM-as-judge protocol. The judge sees two explanations of the same
optimization result: the \emph{verifier-emitted prose subset} of the
grounded narration (i.e., the user-visible output after permissive
emission) and the raw ungrounded narration, without knowing which is
which and without access to the evidence graph. To control for position
bias, every pair is judged twice with the order swapped.

A pair score is computed by averaging the two ordering-wise grounded scores,
producing one of \{1.0, 0.75, 0.5, 0.25, 0.0\} plus an
\textit{inconsistent} flag for pairs where the judge flipped its verdict
across orderings.

Across a deterministic stratified slice of 150 matched pairs (50 per
narrator $\times$ 3 narrators, the first 150 of 450 in lexicographic
(profileId, goal, modelId) order) = 300 judge calls, we observe:

\begin{itemize}
    \item \textbf{0\% tie / parse-failure rate.} Every call returned a
    definitive A or B vote with mean confidence 0.91.
    \item \textbf{83\% observed position-A rate.} Across the 300
    counterbalanced calls, the judge picks the explanation in slot A
    83\% of the time.
    \item \textbf{Mean preference score < 0.5 on all three narrators.}
    Sonnet 0.30, mini 0.41, Llama 0.42: the judge prefers the
    \textit{ungrounded} narration across the board (a usefulness
    preference, not a faithfulness measurement; the judge sees no
    evidence graph).
\end{itemize}

The 83\% call-level position-A vote rate and 66\% pair-level flip rate
measure different quantities and are not algebraically interchangeable.
Together they show a strong position effect while leaving room for
content-sensitive decisions. Within the 34\% of pairs where the judge
agrees across orderings, ungrounded wins decisively.

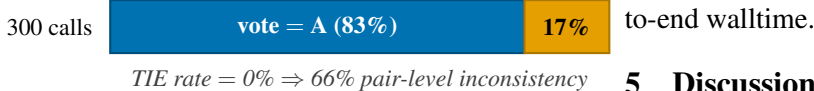
\begin{figure}[t]
\centering
\begin{tikzpicture}[font=\small]
  \def\barwidth{6.6cm}
  \def\barheight{0.7cm}
  \fill[cbBlue, draw=cbBlue!70!black, thick] (0,0) rectangle (0.83*\barwidth, \barheight);
  \fill[cbOrange, draw=cbOrange!75!black, thick] (0.83*\barwidth, 0) rectangle (\barwidth, \barheight);
  \node[white, font=\bfseries\small] at (0.415*\barwidth, 0.5*\barheight) {vote $=$ A (83\%)};
  \node[black, font=\bfseries\small] at (0.915*\barwidth, 0.5*\barheight) {17\%};
  \node[anchor=east, font=\footnotesize] at (-0.1, 0.5*\barheight) {300 calls};
  \node[anchor=north, font=\footnotesize\itshape, text=gray!50!black] at (0.5*\barwidth, -0.08) {TIE rate $=$ 0\%  $\Rightarrow$  66\% pair-level inconsistency};
\end{tikzpicture}

\vspace{0.6em}

\begin{tikzpicture}
\begin{axis}[
  verity axis,
  ybar,
  bar width=24pt,
  width=\linewidth,
  height=4cm,
  ylabel={Mean preference score\\(lower $=$ ungrounded preferred)},
  ylabel style={align=center, font=\footnotesize},
  tick label style={font=\footnotesize},
  symbolic x coords={Sonnet, Llama-70B, mini},
  xtick=data,
  enlarge x limits=0.32,
  ymin=0, ymax=1.05,
  ytick={0, 0.25, 0.5, 0.75, 1.0},
  yticklabels={0.00, 0.25, 0.50, 0.75, 1.00},
  nodes near coords={\pgfmathprintnumber[fixed, precision=2]{\pgfplotspointmeta}},
  nodes near coords style={font=\footnotesize\bfseries, yshift=4pt, anchor=south},
  clip=false,
]
\addplot[draw=cbPurple, fill=cbPurple!30]
  coordinates {(Sonnet, 0.30) (Llama-70B, 0.42) (mini, 0.41)};
\draw[cbVermillion, very thick, dashed] (axis cs:Sonnet, 0.5) -- (axis cs:mini, 0.5);
\node[font=\footnotesize\bfseries, text=cbVermillion, anchor=south,
      fill=white, fill opacity=0.9, text opacity=1, inner sep=2pt]
  at (axis cs:Sonnet, 0.78) {neutral line at $0.5$ $\rightarrow$};
\end{axis}
\end{tikzpicture}
\caption{Counterbalanced judge diagnostic. \textit{Top:} Position A is
chosen in 83\% of 300 calls and 66\% of swapped pairs flip---a strong
position effect, not proof of content independence. \textit{Bottom:} All
three mean scores are below 0.5, favoring raw ungrounded prose. Because
the judge has no evidence graph, this is usefulness preference rather
than faithfulness.}
\label{fig:judge}
\end{figure}

Figure~\ref{fig:judge} visualizes the position-bias diagnostic and per-model
preference distribution side by side.

This is the strongest argument for deterministic verification. A frontier
LLM-as-judge, given full freedom to compare two explanations, picks the
fluent ungrounded one (a preference signal, not faithfulness; no
evidence access). Without counterbalancing, a one-shot judge would
report ``ungrounded wins 83 to 17'': a fake headline driven by position
bias. Counterbalancing exposes the position effect and isolates the
order-consistent subset; we do not interpret the remaining preference as
content-independent.

\paragraph{External rater validation.}
On a 25-item Gates~1--3 sample, three independent raters reach 74\%
mean pairwise agreement and expose structural--prose mismatches. A
separate 25-item Gate~4 study (auxiliary seed 43) finds 88\%
majority-of-three agreement with the strict rule. Appendices
\ref{app:rater-validation}--\ref{app:rater-rubric} give both protocols.

\subsection{Cost and deployment characteristics}
\label{sec:cost}

\noindent\textbf{Cost estimates.} Author-recorded estimates based on run token
counts and vendor list prices on 2026-05-14 give per-explanation cost at
$r{=}1$ of:
\textsc{gpt-4o-mini} \textbf{\$0.001}, \textsc{Llama-3.3-70B}
\textbf{\$0.0016}, \textsc{Claude Sonnet 4.6} \textbf{\$0.05}. Relative
to Llama at $r{=}1$, Sonnet costs $\approx31\times$ as much
(\$0.05/\$0.0016) and ships $\approx1.86\times$ as many verified claims
per instance (6.67/3.58). The final estimates are archived, but the underlying
provider token-count export was not preserved, so they are not independently
re-estimable from the supplement.

\noindent\textbf{Verifier microbench.} On one CPU core using Java~21, a representative
14-claim instance takes p50/p95/p99 $\approx$ 9/16/29\,$\mu$s (10k warmup,
100k measured iterations), sustaining $\approx$88{,}000
verifications/sec/core; observed post-run heap usage was 455.2\,MB.
LLM latency therefore
dominates end-to-end walltime.

\section{Discussion and Lessons Learned}
\label{sec:discussion}

\paragraph{Gate~4 isolated.}
On Sonnet at $r{=}1$, Gates 1--3 fall to 0/0/1 errors while Gate~4 has
841 distinct failing claims; a second pass has no clear effect. Across the
tested tiers this suggests a type-routing bottleneck, not a universal ceiling.

\paragraph{Sparse-explanation edge case.}
At $r{=}1$ grounded, zero-verified-claim rates span \textbf{24.7\%}
(mini) to \textbf{1.3\%} (Sonnet). Deployments need a minimum threshold,
deterministic template fallback, or stronger narrator.

\paragraph{Generalization sketch.}
Graph builders and Gate~4 rules are domain-specific. Medical and loan-risk
mappings are possible examples, not tested transfer.

\section{Related Work}

\paragraph{Structured-data and citation faithfulness.}
Slot-error methods check missing or spurious meaning-representation values
\cite{wen-etal-2015-semantically,dusek-jurcicek-2016-sequence,juraska-etal-2018-deep,nie-etal-2019-simple,dusek-etal-2019-semantic};
table-to-text metrics use extraction or reference alignment
\cite{wiseman-etal-2017-challenges,dhingra-etal-2019-handling}; and ALCE
checks citations~\cite{gao-etal-2023-enabling}. \textsc{VerityGate} adds a
typed declaration/evidence contract and paired repair/filter evaluation,
not a new set-membership operation (Appendix~\ref{app:lineage}).

\paragraph{Other defenses and metrics.}
NLI and atomic-fact metrics cover less structured evidence
\cite{ragas2024,attributable2023,min2023factscore,gao2023rarr,yue2023attrscore,chen2024rgb}.
Complementary work studies constrained syntax~\cite{outlines2023}, repair
\cite{madaan2023selfrefine,yan2024crag,huang2024selfcorrect}, judge bias
\cite{zheng2023judging,wang2024positionbias}, and tool safety
\cite{mcpsafety2025,greshake2023}.

\section*{Ethical Considerations}

This work studies hallucination in LLM-generated explanations of structured
decisions. All experiments were conducted in controlled local environments
against a credit card portfolio benchmark. No production systems or
production end users were involved. Three external raters provided offline
labels for a small sample of generated claims; the released rater files
contain no names or directly identifying information. No directly identifying
personal data was processed: the benchmark profiles used in this paper are generated
parametrically; prior backend validation referenced in Section~\ref{sec:eval}
used \emph{public-use} BLS microdata, not personally identifiable
records. We believe that documenting
the failure modes of grounded LLM narration (and providing a deployable
defense) serves the public interest, since organizations cannot protect
against risks they do not understand. \textbf{Risk of false assurance
from partial verification.} A ``verified'' label produced by
\textsc{VerityGate} signals \emph{schema-level} consistency between the
declared claim fields and the evidence graph; it does \emph{not}
warrant that the prose shown to the user is faithful in full
(see Appendix~\ref{app:rater-validation}). Deployments must surface
this scope to end users (e.g., via product copy and/or audit logs) and
must not present the label as a guarantee of complete factual
accuracy. The \veritycoderelease{} provides the framework implementation;
the prompts, schemas, pre-registration, and reported per-cell logs are
archived in the \verityartifact{} under the repository's permissive license.

\section*{Limitations}

\textsc{VerityGate} has several limitations.

First, we evaluate against three LLM backends. While these span a wide
capability range (small / open-weights mid / frontier), the rapidly evolving
landscape means results may not generalize to all future models. The
observed pattern (repair helps the small model most while Gate~4 remains
dominant in all three tested tiers) may not hold for other models; both
the pattern and absolute numbers require replication.

Second, we use a single seed (42) for profile generation. We report tight
paired bootstrap CIs as the primary uncertainty quantification rather than
running multiple seeds. This is sufficient for the within-bench claims but
does not assess seed robustness.

Third, our external rater studies (Appendix~\ref{app:rater-validation})
both use small stratified samples (3 raters $\times$ 25 items each, for
Gates 1--3 and for Gate~4 separately). Larger studies with
domain-expert raters and broader coverage are a clear follow-up. Note
that the verifier's primary check is against the deterministic
backend's allowlists, not against human judgment; the rater studies
probe a different question (\textit{do humans, given the same
allowlists, agree with the verifier's accept/reject calls?}) rather
than re-deriving the ground truth.

Fourth, the LLM-as-judge analysis was conducted using a single judge model
(\textsc{gpt-4o}). Other judge models may exhibit different position-bias
magnitudes and different absolute preferences. We report the judge result
as a secondary metric and explicitly tag it as ``blinded pairwise
usefulness preference under counterbalancing,'' not as a faithfulness
measure.

Fifth, adversarial robustness is not evaluated. The verifier is sound
against non-adversarial LLM hallucination, but a model that knows the
allowlist could craft outputs that pass verification while remaining
semantically misleading. This is out of scope for this paper, since our
threat model is unintentional fabrication, not adversarial attack.

Sixth, all empirical results are on a single application domain
(credit-card portfolio optimization with the MILP backend described in
\S\ref{sec:eval}). The verifier interface, claim schema, and Gate~4
type rules are designed to be portable, and we sketch mappings to
medical decision support and loan-risk assessment in
Section~\ref{sec:discussion}; neither mapping is empirically tested.
Empirical cross-domain transfer is the most important follow-up.

Seventh, we do not compare narrative quality or coverage against a
deterministic evidence-graph template, ablate strict versus relaxed
Gate~4 rules, or run NLI/RAGAS/ALCE scorers on the generated outputs.
The results therefore establish the behavior of the proposed structural
audit-and-filter contract, not superiority over template generation or
complementarity to neural faithfulness metrics.

\bibliography{references}

@article{feinstein1990,
  author  = {Feinstein, Alvan R. and Cicchetti, Domenic V.},
  title   = {High agreement but low kappa: I. The problems of two paradoxes},
  journal = {Journal of Clinical Epidemiology},
  volume  = {43},
  number  = {6},
  pages   = {543--549},
  year    = {1990},
  doi     = {10.1016/0895-4356(90)90158-L}
}

@article{gwet2008,
  author  = {Gwet, Kilem L.},
  title   = {Computing inter-rater reliability and its variance in the presence of high agreement},
  journal = {British Journal of Mathematical and Statistical Psychology},
  volume  = {61},
  number  = {1},
  pages   = {29--48},
  month   = may,
  year    = {2008},
  doi     = {10.1348/000711006X126600}
}

@article{landis1977,
  author  = {Landis, J. Richard and Koch, Gary G.},
  title   = {The Measurement of Observer Agreement for Categorical Data},
  journal = {Biometrics},
  volume  = {33},
  number  = {1},
  pages   = {159--174},
  year    = {1977},
  doi     = {10.2307/2529310}
}

@misc{bls_cex_pumd,
  author       = {{U.S. Bureau of Labor Statistics}},
  title        = {Consumer Expenditure Surveys ({CE}), Interview Survey: Public-Use Microdata ({PUMD}), 2020--2024},
  year         = {2024},
  howpublished = {\url{https://www.bls.gov/cex/pumd.htm}},
  note         = {Accessed January 2026}
}

@inproceedings{ragas2024,
  title     = {{RAGA}s: Automated Evaluation of Retrieval Augmented Generation},
  author    = {Es, Shahul and James, Jithin and Espinosa Anke, Luis and Schockaert, Steven},
  booktitle = {Proceedings of the 18th Conference of the European Chapter of the Association for Computational Linguistics: System Demonstrations},
  month     = mar,
  pages     = {150--158},
  year      = {2024},
  address   = {St. Julians, Malta},
  publisher = {Association for Computational Linguistics},
  url       = {https://aclanthology.org/2024.eacl-demo.16/},
  doi       = {10.18653/v1/2024.eacl-demo.16}
}

@article{attributable2023,
  title   = {Measuring Attribution in Natural Language Generation Models},
  author  = {Rashkin, Hannah and Nikolaev, Vitaly and Lamm, Matthew and Aroyo, Lora and Collins, Michael and Das, Dipanjan and Petrov, Slav and Tomar, Gaurav Singh and Turc, Iulia and Reitter, David},
  journal = {Computational Linguistics},
  volume  = {49},
  number  = {4},
  pages   = {777--840},
  year    = {2023},
  url     = {https://aclanthology.org/2023.cl-4.2/},
  doi     = {10.1162/coli_a_00486}
}

@misc{outlines2023,
  title   = {Efficient Guided Generation for Large Language Models},
  author  = {Willard, Brandon T. and Louf, R{\'e}mi},
  year    = {2023},
  eprint  = {2307.09702},
  archivePrefix = {arXiv},
  primaryClass = {cs.CL},
  url     = {https://arxiv.org/abs/2307.09702}
}

@inproceedings{madaan2023selfrefine,
  title     = {Self-Refine: Iterative Refinement with Self-Feedback},
  author    = {Madaan, Aman and Tandon, Niket and Gupta, Prakhar and Hallinan, Skyler and Gao, Luyu and Wiegreffe, Sarah and Alon, Uri and Dziri, Nouha and Prabhumoye, Shrimai and Yang, Yiming and Gupta, Shashank and Majumder, Bodhisattwa Prasad and Hermann, Katherine and Welleck, Sean and Yazdanbakhsh, Amir and Clark, Peter},
  booktitle = {Advances in Neural Information Processing Systems},
  volume    = {36},
  pages     = {46534--46594},
  publisher = {Curran Associates, Inc.},
  year      = {2023},
  url       = {https://proceedings.neurips.cc/paper_files/paper/2023/hash/91edff07232fb1b55a505a9e9f6c0ff3-Abstract-Conference.html},
  doi       = {10.52202/075280-2019}
}

@inproceedings{huang2024selfcorrect,
  title     = {Large Language Models Cannot Self-Correct Reasoning Yet},
  author    = {Huang, Jie and Chen, Xinyun and Mishra, Swaroop and Zheng, Huaixiu Steven and Yu, Adams Wei and Song, Xinying and Zhou, Denny},
  booktitle = {The Twelfth International Conference on Learning Representations},
  year      = {2024},
  url       = {https://openreview.net/forum?id=IkmD3fKBPQ}
}

@inproceedings{zheng2023judging,
  title     = {Judging {LLM}-as-a-Judge with {MT-Bench} and Chatbot Arena},
  author    = {Zheng, Lianmin and Chiang, Wei-Lin and Sheng, Ying and Zhuang, Siyuan and Wu, Zhanghao and Zhuang, Yonghao and Lin, Zi and Li, Zhuohan and Li, Dacheng and Xing, Eric and Zhang, Hao and Gonzalez, Joseph and Stoica, Ion},
  booktitle = {Advances in Neural Information Processing Systems},
  volume    = {36},
  pages     = {46595--46623},
  publisher = {Curran Associates, Inc.},
  year      = {2023},
  url       = {https://proceedings.neurips.cc/paper_files/paper/2023/hash/91f18a1287b398d378ef22505bf41832-Abstract-Datasets_and_Benchmarks.html},
  doi       = {10.52202/075280-2020}
}

@inproceedings{wang2024positionbias,
  title     = {Large Language Models are not Fair Evaluators},
  author    = {Wang, Peiyi and Li, Lei and Chen, Liang and Cai, Zefan and Zhu, Dawei and Lin, Binghuai and Cao, Yunbo and Kong, Lingpeng and Liu, Qi and Liu, Tianyu and Sui, Zhifang},
  booktitle = {Proceedings of the 62nd Annual Meeting of the Association for Computational Linguistics (Volume 1: Long Papers)},
  month     = aug,
  pages     = {9440--9450},
  year      = {2024},
  address   = {Bangkok, Thailand},
  publisher = {Association for Computational Linguistics},
  url       = {https://aclanthology.org/2024.acl-long.511/},
  doi       = {10.18653/v1/2024.acl-long.511}
}

@article{mcpsafety2025,
  title   = {{MCP} Safety Audit: {LLM}s with the Model Context Protocol allow major security exploits},
  author  = {Radosevich, Brandon and Halloran, John},
  journal = {arXiv preprint arXiv:2504.03767},
  year    = {2025},
  url     = {https://arxiv.org/abs/2504.03767}
}

@inproceedings{greshake2023,
  title     = {Not what you've signed up for: Compromising real-world {LLM}-integrated applications with indirect prompt injection},
  author    = {Greshake, Kai and Abdelnabi, Sahar and Mishra, Shailesh and Endres, Christoph and Holz, Thorsten and Fritz, Mario},
  booktitle = {Proceedings of the 16th ACM Workshop on Artificial Intelligence and Security},
  pages     = {79--90},
  publisher = {Association for Computing Machinery},
  address   = {New York, NY, USA},
  location  = {Copenhagen, Denmark},
  series    = {{AISec} '23},
  year      = {2023},
  url       = {https://doi.org/10.1145/3605764.3623985},
  doi       = {10.1145/3605764.3623985}
}

@misc{moffatt2024,
  title        = {Moffatt v. Air Canada, 2024 BCCRT 149},
  author       = {{British Columbia Civil Resolution Tribunal}},
  year         = {2024},
  month        = feb,
  howpublished = {\url{https://www.canlii.org/en/bc/bccrt/doc/2024/2024bccrt149/2024bccrt149.html}},
  note         = {Decision dated February 14, 2024. Accessed May 2026.}
}

@misc{nyc_mycity_2024,
  title        = {{NYC's} {AI} Chatbot Tells Businesses to Break the Law},
  author       = {Lecher, Colin},
  howpublished = {The Markup},
  year         = {2024},
  month        = mar,
  url          = {https://themarkup.org/artificial-intelligence/2024/03/29/nycs-ai-chatbot-tells-businesses-to-break-the-law},
  note         = {Accessed May 2026.}
}

@misc{morgan_morgan_2025,
  title        = {Morgan \& Morgan Lawyers Fined for ``Hallucinated'' {AI} Citations},
  author       = {Pazanowski, Bernie},
  howpublished = {Bloomberg Law},
  year         = {2025},
  month        = feb,
  url          = {https://news.bloomberglaw.com/us-law-week/morgan-morgan-lawyers-fined-for-hallucinated-ai-citations},
  note         = {Wadsworth v. Walmart, U.S. District Court for the District of Wyoming; Judge Kelly H. Rankin sanctions order dated February 24, 2025. Accessed May 2026.}
}

@misc{charlotin_hallucination_db,
  title        = {{AI} Hallucination Cases Database},
  author       = {Charlotin, Damien},
  howpublished = {\url{https://www.damiencharlotin.com/hallucinations/}},
  year         = {2025},
  note         = {Database is updated continuously; figure cited (over 1{,}400 rulings) reflects a May 2026 snapshot. Accessed May 2026.}
}

@inproceedings{min2023factscore,
  title     = {{F}{A}ct{S}core: Fine-grained Atomic Evaluation of Factual Precision in Long Form Text Generation},
  author    = {Min, Sewon and Krishna, Kalpesh and Lyu, Xinxi and Lewis, Mike and Yih, Wen-tau and Koh, Pang and Iyyer, Mohit and Zettlemoyer, Luke and Hajishirzi, Hannaneh},
  booktitle = {Proceedings of the 2023 Conference on Empirical Methods in Natural Language Processing},
  month     = dec,
  pages     = {12076--12100},
  year      = {2023},
  address   = {Singapore},
  publisher = {Association for Computational Linguistics},
  url       = {https://aclanthology.org/2023.emnlp-main.741/},
  doi       = {10.18653/v1/2023.emnlp-main.741}
}

@inproceedings{gao2023rarr,
  title     = {{RARR}: Researching and Revising What Language Models Say, Using Language Models},
  author    = {Gao, Luyu and Dai, Zhuyun and Pasupat, Panupong and Chen, Anthony and Chaganty, Arun Tejasvi and Fan, Yicheng and Zhao, Vincent and Lao, Ni and Lee, Hongrae and Juan, Da-Cheng and Guu, Kelvin},
  booktitle = {Proceedings of the 61st Annual Meeting of the Association for Computational Linguistics (Volume 1: Long Papers)},
  month     = jul,
  pages     = {16477--16508},
  year      = {2023},
  address   = {Toronto, Canada},
  publisher = {Association for Computational Linguistics},
  url       = {https://aclanthology.org/2023.acl-long.910/},
  doi       = {10.18653/v1/2023.acl-long.910}
}

@article{yan2024crag,
  title   = {Corrective Retrieval Augmented Generation},
  author  = {Yan, Shi-Qi and Gu, Jia-Chen and Zhu, Yun and Ling, Zhen-Hua},
  journal = {arXiv preprint arXiv:2401.15884},
  year    = {2024},
  url     = {https://arxiv.org/abs/2401.15884}
}

@inproceedings{yue2023attrscore,
  title     = {Automatic Evaluation of Attribution by Large Language Models},
  author    = {Yue, Xiang and Wang, Boshi and Chen, Ziru and Zhang, Kai and Su, Yu and Sun, Huan},
  booktitle = {Findings of the Association for Computational Linguistics: EMNLP 2023},
  month     = dec,
  pages     = {4615--4635},
  year      = {2023},
  address   = {Singapore},
  publisher = {Association for Computational Linguistics},
  url       = {https://aclanthology.org/2023.findings-emnlp.307/},
  doi       = {10.18653/v1/2023.findings-emnlp.307}
}

@inproceedings{chen2024rgb,
  title     = {Benchmarking Large Language Models in Retrieval-Augmented Generation},
  author    = {Chen, Jiawei and Lin, Hongyu and Han, Xianpei and Sun, Le},
  booktitle = {Proceedings of the {AAAI} Conference on Artificial Intelligence},
  volume    = {38},
  pages     = {17754--17762},
  year      = {2024},
  url       = {https://ojs.aaai.org/index.php/AAAI/article/view/29728},
  doi       = {10.1609/aaai.v38i16.29728}
}

@inproceedings{wen-etal-2015-semantically,
  title     = {Semantically Conditioned {LSTM}-based Natural Language Generation for Spoken Dialogue Systems},
  author    = {Wen, Tsung-Hsien and Ga{\v{s}}i{\'c}, Milica and Mrk{\v{s}}i{\'c}, Nikola and Su, Pei-Hao and Vandyke, David and Young, Steve},
  booktitle = {Proceedings of the 2015 Conference on Empirical Methods in Natural Language Processing},
  month     = sep,
  year      = {2015},
  address   = {Lisbon, Portugal},
  publisher = {Association for Computational Linguistics},
  url       = {https://aclanthology.org/D15-1199/},
  doi       = {10.18653/v1/D15-1199},
  pages     = {1711--1721}
}

@inproceedings{dusek-jurcicek-2016-sequence,
  title     = {Sequence-to-Sequence Generation for Spoken Dialogue via Deep Syntax Trees and Strings},
  author    = {Du{\v{s}}ek, Ond{\v{r}}ej and Jur{\v{c}}{\'i}{\v{c}}ek, Filip},
  booktitle = {Proceedings of the 54th Annual Meeting of the Association for Computational Linguistics (Volume 2: Short Papers)},
  month     = aug,
  year      = {2016},
  address   = {Berlin, Germany},
  publisher = {Association for Computational Linguistics},
  url       = {https://aclanthology.org/P16-2008/},
  doi       = {10.18653/v1/P16-2008},
  pages     = {45--51}
}

@inproceedings{juraska-etal-2018-deep,
  title     = {A Deep Ensemble Model with Slot Alignment for Sequence-to-Sequence Natural Language Generation},
  author    = {Juraska, Juraj and Karagiannis, Panagiotis and Bowden, Kevin and Walker, Marilyn},
  booktitle = {Proceedings of the 2018 Conference of the North {A}merican Chapter of the Association for Computational Linguistics: Human Language Technologies, Volume 1 (Long Papers)},
  month     = jun,
  year      = {2018},
  address   = {New Orleans, Louisiana},
  publisher = {Association for Computational Linguistics},
  url       = {https://aclanthology.org/N18-1014/},
  doi       = {10.18653/v1/N18-1014},
  pages     = {152--162}
}

@inproceedings{nie-etal-2019-simple,
  title     = {A Simple Recipe towards Reducing Hallucination in Neural Surface Realisation},
  author    = {Nie, Feng and Yao, Jin-Ge and Wang, Jinpeng and Pan, Rong and Lin, Chin-Yew},
  booktitle = {Proceedings of the 57th Annual Meeting of the Association for Computational Linguistics},
  month     = jul,
  year      = {2019},
  address   = {Florence, Italy},
  publisher = {Association for Computational Linguistics},
  url       = {https://aclanthology.org/P19-1256/},
  doi       = {10.18653/v1/P19-1256},
  pages     = {2673--2679}
}

@inproceedings{dusek-etal-2019-semantic,
  title     = {Semantic Noise Matters for Neural Natural Language Generation},
  author    = {Du{\v{s}}ek, Ond{\v{r}}ej and Howcroft, David M. and Rieser, Verena},
  booktitle = {Proceedings of the 12th International Conference on Natural Language Generation},
  month     = oct # {--} # nov,
  year      = {2019},
  address   = {Tokyo, Japan},
  publisher = {Association for Computational Linguistics},
  url       = {https://aclanthology.org/W19-8652/},
  doi       = {10.18653/v1/W19-8652},
  pages     = {421--426}
}

@inproceedings{wiseman-etal-2017-challenges,
  title     = {Challenges in Data-to-Document Generation},
  author    = {Wiseman, Sam and Shieber, Stuart and Rush, Alexander},
  booktitle = {Proceedings of the 2017 Conference on Empirical Methods in Natural Language Processing},
  month     = sep,
  year      = {2017},
  address   = {Copenhagen, Denmark},
  publisher = {Association for Computational Linguistics},
  url       = {https://aclanthology.org/D17-1239/},
  doi       = {10.18653/v1/D17-1239},
  pages     = {2253--2263}
}

@inproceedings{dhingra-etal-2019-handling,
  title     = {Handling Divergent Reference Texts when Evaluating Table-to-Text Generation},
  author    = {Dhingra, Bhuwan and Faruqui, Manaal and Parikh, Ankur and Chang, Ming-Wei and Das, Dipanjan and Cohen, William},
  booktitle = {Proceedings of the 57th Annual Meeting of the Association for Computational Linguistics},
  month     = jul,
  year      = {2019},
  address   = {Florence, Italy},
  publisher = {Association for Computational Linguistics},
  url       = {https://aclanthology.org/P19-1483/},
  doi       = {10.18653/v1/P19-1483},
  pages     = {4884--4895}
}

@inproceedings{gao-etal-2023-enabling,
  title     = {Enabling Large Language Models to Generate Text with Citations},
  author    = {Gao, Tianyu and Yen, Howard and Yu, Jiatong and Chen, Danqi},
  booktitle = {Proceedings of the 2023 Conference on Empirical Methods in Natural Language Processing},
  month     = dec,
  year      = {2023},
  address   = {Singapore},
  publisher = {Association for Computational Linguistics},
  url       = {https://aclanthology.org/2023.emnlp-main.398/},
  doi       = {10.18653/v1/2023.emnlp-main.398},
  pages     = {6465--6488}
}

\appendix
\refstepcounter{section}\label{app:rater-validation}\section*{Appendix~\Alph{section}. Independent Rater Validation}

Three independent raters with computer science backgrounds (no specific
ML training) each rated the same stratified 25-item sample drawn from
our paper-grade evaluation run: 10 items the verifier classified as
VERIFIED and 15 items rejected by Gates 1--3. Items rejected by Gate~4
were excluded from the rater protocol because that gate requires
domain-specific structural judgment beyond the rubric's scope; we
discuss Gate~4 separately in Section~\ref{sec:discussion}.

Raters followed a fixed rubric (Appendix~\ref{app:rater-rubric}) that
asked a single binary question per item: \textit{Does this claim
reference only entities and numbers that appear in the evidence box?}
The rubric was designed to require careful reading and table
comprehension only, matching the mechanical nature of Gates 1--3.
Raters had no visibility into the verifier's classification when
rating. Items rated UNCLEAR by any rater were excluded from
agreement calculations following standard practice.

Mean pairwise agreement across the three raters was 74\% on the
25-item sample. Table~\ref{tab:rater-agreement} reports the full chance-corrected agreement statistics: pairwise Cohen's $\kappa$, Fleiss' $\kappa$ across all three raters, and each rater's agreement with the verifier as the reference label. Fleiss' $\kappa = 0.48$ indicates moderate inter-rater agreement~\cite{landis1977}; pairwise Cohen's $\kappa$ ranges from $0.35$ (substantively fair) to $0.70$ (substantial), with the wide span itself
informative: rubric-following raters can disagree even on a deterministic structural check. Rater-vs-verifier $\kappa$ is markedly lower (0.00--0.30), and the majority-of-3 vs.\ verifier $\kappa$ is 0.02, consistent with the systematic structural-vs-prose disagreement pattern we discuss below. We interpret these numbers as supporting, not undermining, the case for mechanical verification: human raters disagree both with each other and (more sharply) with the verifier, while the verifier is reproducibly self-consistent across runs.

\begin{table}[t]
\centering
\small
\setlength{\tabcolsep}{4pt}
\begin{tabular}{lrrr}
\toprule
\textbf{Comparison} & \textbf{Agreement} & \textbf{Cohen's $\kappa$} & $n$ \\
\midrule
Rater 1 vs Rater 2 & 69.6\% & 0.39 & 23 \\
Rater 1 vs Rater 3 & 66.7\% & 0.35 & 21 \\
Rater 2 vs Rater 3 & 85.7\% & 0.70 & 21 \\
\midrule
Rater 1 vs Verifier & 65.2\% & 0.30 & 23 \\
Rater 2 vs Verifier & 52.0\% & 0.03 & 25 \\
Rater 3 vs Verifier & 42.9\% & 0.00 & 21 \\
Majority-of-3 vs Verifier & 50.0\% & 0.02 & 24 \\
\midrule
Fleiss' $\kappa$ (3 raters) & --- & \textbf{0.48} & 21 \\
\bottomrule
\end{tabular}
\caption{Chance-corrected inter-rater and rater-vs-verifier agreement on
the 25-item rater sample. UNCLEAR votes dropped from each comparison;
sample sizes reflect per-pair retention. Fleiss' $\kappa$ uses items where
all three raters voted YES/NO (21 items).}
\label{tab:rater-agreement}
\end{table}

The qualitative disagreement pattern between raters and the verifier
is systematic rather than noise:

\begin{itemize}
  \item \textbf{All three raters disagreed with the verifier on four
    items} (a unanimous \textsc{Yes} vs.\ verifier \textsc{No}). The
    raters treated numerically equivalent trailing-zero forms as the same,
    while the verifier required exact membership of each declared number
    string. For example, \texttt{2400.0} did not match the allowed
    \texttt{2400.00}; one of the four items also failed Gate~4. This
    exposes an exact-string canonicalization limitation.
  \item \textbf{All three raters disagreed with the verifier in the
    opposite direction on one item} (unanimous \textsc{No} vs.\ verifier
    \textsc{Yes}). The claim's prose mentioned a phrase not in the
    allowed-entity list (``general purchases''), but the LLM emitted
    empty cited fields and the verifier had nothing to check.
\end{itemize}

These five items (4 + 1) account for the majority of the gap between
rater majority vote and the verifier. They argue for numeric
canonicalization and combined \textit{structural + prose-level}
verification as future work, while
the structural check alone catches the failure classes (Gates 1--4)
with reproducibility that subjective human judgment partially lacks
(3 raters do not perfectly agree among themselves; mean pairwise
agreement 74\%).

\paragraph{Walkthrough~1: Verified claim, unanimous agreement (item 4).}
Claim: \textit{``chase-freedom-flex earns \$197.64, credits \$0.00, fee
\$0.00, net \$197.64''}. Allowed entities include
\texttt{chase-freedom-flex}; allowed numbers include $0$ and $197.64$.
Every entity and number in the prose appears in the evidence box.
Verifier: \textsc{Yes}. All three raters: \textsc{Yes}. Clean case
testing the rubric's \textit{strict-literal} criterion.

\paragraph{Walkthrough~2: Gate 3 violation, all raters agree (item 13).}
Claim: \textit{``The Capital One Savor card offers 4\% cashback on
dining, making it ideal for food-related expenses compared to standard
cashback cards.''} Allowed entities include \texttt{cap1-savor} (matches
``Capital One Savor'' by paraphrase) and \texttt{DINING}; allowed
numbers do not include $4$. Verifier: \textsc{No} (Gate~3). All three
raters: \textsc{No} (correctly identified the unallowed ``4''). A clean
Gate~3 violation case where rater majority and verifier converge on
rejection.

\paragraph{Walkthrough~3: Verifier-vs-rater disagreement, structural-only
limitation (item 17).} Claim: \textit{``Segment allocation for ONLINE
shows amazon-prime-visa at \$2400.00 and amex-bcp at \$6000.00.''}
Allowed entities include \texttt{ONLINE}, \texttt{amazon-prime-visa},
\texttt{amex-bcp}; allowed numbers include $2400$ and $6000$. Every
prose element appears in the displayed evidence box by inspection. All three raters:
\textsc{Yes}. Verifier: \textsc{No}. The declared numbers were
\texttt{2400.0} and \texttt{6000.0}, whereas the allowed strings were
\texttt{2400.00} and \texttt{6000.00}. The prose values are numerically
correct, but the evaluated exact-string check rejects the structured
fields. This case motivates canonical numeric comparison in a future
verifier version.

\paragraph{Gate~4 validation sub-study.}
The 25-item study above by design covers Gates~1--3 only. Because
Gate~4 dominates residual failures, we ran a separate three-rater
sub-study over 25 stratified Gate~4 failures sampled from a
re-instrumented smoke run (7 \texttt{COMPARISON} + 6
\texttt{FEE\_JUSTIFICATION} + 6 \texttt{CAP\_SWITCH} + 6
\texttt{ASSUMPTION}, round-robin across the three narrators, seed 43).
Raters saw the claim text and the human-readable types of the evidence
records the LLM cited; the verifier's expected-type column and decision
were withheld to avoid priming. Raters judged whether the cited evidence
type(s) were a reasonable backing for the claim text (\textsc{Agree})
or not (\textsc{Disagree}); \textsc{Unclear} was a third option.

Table~\ref{tab:gate4-rater} reports the agreement statistics.
\textbf{Majority-of-3 raters agreed with the verifier's rejection on
22 of 25 items (88\%);} mean per-rater agreement with the verifier
was 86\% (92\%, 71\%, 96\%). Pairwise Cohen's $\kappa$ ranges 0.19--0.36
and Fleiss' $\kappa$ across the three raters is 0.19, but these
chance-corrected values are noise-dominated under the highly skewed
prevalence (most items are clearly verifier-correct rejections):
when $p_e \approx p_o$, $\kappa$ collapses irrespective of the
substantive level of agreement~\cite{feinstein1990,gwet2008}.
$\kappa$ against the verifier reference is undefined by construction
(the verifier rejected every item, giving zero variance) and is
reported here as $0.00$ for completeness. Raw \% agreement is the
substantive metric under this prevalence regime.

\begin{table}[h]
\centering
\small
\setlength{\tabcolsep}{4pt}
\begin{tabular}{lrrr}
\toprule
\textbf{Comparison} & \textbf{Agreement} & \textbf{Cohen's $\kappa$} & $n$ \\
\midrule
Rater 1 vs Rater 2 & 79.2\% & 0.36 & 24 \\
Rater 1 vs Rater 3 & 88.0\% & $-$0.06 & 25 \\
Rater 2 vs Rater 3 & 75.0\% & 0.19 & 24 \\
\midrule
Rater 1 vs Verifier & 92.0\% & 0.00 & 25 \\
Rater 2 vs Verifier & 70.8\% & 0.00 & 24 \\
Rater 3 vs Verifier & 96.0\% & 0.00 & 25 \\
\textbf{Majority-of-3 vs Verifier} & \textbf{88.0\%} & 0.00 & 25 \\
\midrule
Fleiss' $\kappa$ (3 raters) & --- & 0.19 & 24 \\
\bottomrule
\end{tabular}
\caption{Gate~4 validation sub-study: rater agreement on 25 stratified
Gate~4 failures, blind rubric (verifier's expected type and decision
hidden). $\kappa$ values are deflated by the prevalence-skew paradox
(verifier rejected every item, raters mostly agreed) and are reported
alongside raw agreement, which is the substantive metric under skewed
marginals. Majority-of-3 vs verifier 88\% is the headline.}
\label{tab:gate4-rater}
\end{table}

The single rater with the lowest verifier agreement (Rater~2, 71\%)
showed a coherent pattern: they accepted \texttt{RESULT\_BREAKDOWN}
(final-money-totals records) as adequate backing for
\texttt{COMPARISON} claims (e.g., \textit{``compares result
breakdown''}). This is a defensible \emph{looser-grounding}
interpretation in which related-but-not-exact evidence types are
acceptable. We report it as a finding rather than noise: Gate~4
enforces a strict type-rule discipline, but a looser standard is
intelligible and an alternative design choice. The fact that two
raters land at $\geq 92$\% verifier agreement and one rater at 71\%
with a coherent rationale gives a useful range, not a sharp validation
ceiling.

\refstepcounter{section}\label{app:rater-rubric}\section*{Appendix~\Alph{section}. Rater Rubric}

Raters received the following one-page protocol prior to evaluation:

\begin{quote}\small
For each paired (claim, evidence) item, answer:
\textit{Does this claim reference only entities and numbers that appear
in the evidence box?}
Choose one of: \textbf{YES} (every entity and number in the claim
appears in the evidence box), \textbf{NO} (at least one entity or
number is not in the evidence box), or \textbf{UNCLEAR} (the claim is
too vague to extract entities/numbers, or an entity name is paraphrased
and you cannot tell if it matches).
\end{quote}

The rubric included five calibration examples covering YES, NO due to
entity violation, NO due to number violation, UNCLEAR, and NO due to a
derived (sum-of-allowed) number. Raters were instructed to be literal
(do not use outside knowledge), to not perform arithmetic (derived
numbers must appear directly in the evidence), and to trust their first
read (mark UNCLEAR rather than debate for more than 60~seconds).

\refstepcounter{section}\label{app:schema}\section*{Appendix~\Alph{section}. Claim Schema and Gate 4 Type Rules}

This appendix gives the six-field prompt/schema contract. The evaluated
parser requires \texttt{claimId}, \texttt{claimType}, and \texttt{text}; it
defaults missing or non-array citation fields to empty arrays. Gates 1--3
inspect those declared arrays, and Gate~4 applies Table~\ref{tab:type-rules}.
The type rules are specific to the credit-card domain.

\paragraph{Claim object (per-claim).}
The prompt asks the LLM to emit a JSON object with six fields:

\begin{quote}\small\ttfamily
\{ \\
\phantom{XX}claimId:\hspace{0.6em}string \\
\phantom{XX}claimType:\hspace{0.6em}enum (one of the six in Table~\ref{tab:type-rules}) \\
\phantom{XX}text:\hspace{0.6em}string \hfill // the user-visible prose \\
\phantom{XX}citedEvidenceIds:\hspace{0.6em}array of string \hfill // Gate 1 input \\
\phantom{XX}citedEntities:\hspace{0.6em}\hspace{0.4em}array of string \hfill // Gate 2 input \\
\phantom{XX}citedNumbers:\hspace{0.6em}\hspace{0.7em}array of string \hfill // Gate 3 input \\
\}
\end{quote}

Gates 1--3 check, respectively, that every element of
\texttt{citedEvidenceIds} is a node in the evidence graph, that every
element of \texttt{citedEntities} is in the per-instance allowed-entity
set, and that every element of \texttt{citedNumbers} exactly matches a
string in the per-instance allowed-number set. Gate~4 applies
\textit{type-specific structural rules}: it requires that at least one
citation in \texttt{citedEvidenceIds} resolve to an evidence node of a
required type. The six claim types and their Gate~4 rules in our
credit-card domain are:

\begin{table}[h]
\centering
\small
\setlength{\tabcolsep}{4pt}
\begin{tabularx}{\columnwidth}{@{}lX@{}}
\toprule
\textbf{Claim type} & \textbf{Required evidence type(s) at Gate 4} \\
\midrule
\texttt{COMPARISON} & must cite \texttt{WINNER\_BY\_CATEGORY} \\
\texttt{ALLOCATION} & (no Gate 4 requirement) \\
\texttt{THRESHOLD} & (no Gate 4 requirement) \\
\texttt{ASSUMPTION} & must cite \texttt{ASSUMPTION} \\
\texttt{FEE\_JUSTIFICATION} & must cite \texttt{FEE\_BREAK\_EVEN} \\
\texttt{CAP\_SWITCH} & must cite \texttt{CAP\_HIT}; when present in the
   graph, must cite both \texttt{CAP\_HIT} and \texttt{ALLOCATION\_SEGMENT} \\
\bottomrule
\end{tabularx}
\caption{Gate~4 (claim-type rules) per claim type. The
``required evidence type'' column lists the evidence-graph node type(s)
that at least one element of \texttt{citedEvidenceIds} must resolve to.
\texttt{CAP\_SWITCH} is the only claim type requiring two distinct
evidence types (one of each) when those types are present in the graph;
the separate unconditional \texttt{CAP\_HIT} path explains the duplicate
message unit described below. The others require at most one.
\texttt{ALLOCATION} and \texttt{THRESHOLD} have no Gate~4 rule; their
declared entities and numbers are checked only by Gates 2 and 3.}
\label{tab:type-rules}
\end{table}

\paragraph{Gate~4 failure modes in practice.}
Table~\ref{tab:gate4-submode} breaks down the residual Gate~4 errors
on the grounded variant at $r{=}1$ by the claim type that triggered
the violation. On every model, \texttt{COMPARISON} (missing
\texttt{WINNER\_BY\_CATEGORY} citation) and \texttt{FEE\_JUSTIFICATION}
(missing \texttt{FEE\_BREAK\_EVEN} citation) are the largest two
sub-modes. \texttt{CAP\_SWITCH} is a smaller fraction in absolute count
but distinctive in its sub-structure. Sonnet produces 220 verifier
messages from 106 distinct \texttt{CAP\_SWITCH} claims. At claim level,
89 (84\%) omit \texttt{CAP\_HIT}, 72 (68\%) omit
\texttt{ALLOCATION\_SEGMENT}, and 55 (52\%) omit both. The last figure
directly supports the interpretation that the model does not
consistently recognize that \texttt{CAP\_SWITCH} requires both evidence
types. At message level, the 220 total comprises 89 messages from the
unconditional required-type check, 59 additional \texttt{CAP\_HIT}
messages from a guarded two-type check, and 72
\texttt{ALLOCATION\_SEGMENT} messages. Thus any third message on a claim
reflects the two verifier paths for \texttt{CAP\_HIT}, not a data or
aggregation error. Constrained decoding over the claim-type schema is the
natural next intervention.

\paragraph{Gate~4 is predominantly type-routing.}
A re-instrumented smoke replay (3 profiles $\times$ 3 goals $\times$ 3
narrators at $r{=}1$, with per-claim cited-evidence type capture)
yielded 139 grounded Gate~4 failures with full citation records.
\textbf{Within this 27-instance replay, 138/139 (99.3\%) had non-empty
\texttt{citedEvidenceIds}}: nearly every rejected claim cited something,
but not the required type; one was true under-citation. The dominant
mis-routing patterns were:
\texttt{FEE\_JUSTIFICATION} claims cited
\texttt{WINNER\_BY\_CATEGORY} in 62.5\% of their failures (25 / 40);
\texttt{COMPARISON} cited \texttt{FEE\_BREAK\_EVEN} in 30.1\% (22 / 73);
\texttt{ASSUMPTION} cited \texttt{WINNER\_BY\_CATEGORY} in 55.6\%
(10 / 18). The required evidence type was available in the graph in
the great majority of cases (verified by sibling claims of the same
type passing Gate~4 within the same instance, e.g., a
\texttt{FEE\_JUSTIFICATION} that correctly cited
\texttt{FEE\_BREAK\_EVEN}). The prompt provides the evidence graph
with each node's type in JSON but does not enumerate the
type$\to$required-evidence mapping; even the repair pass, which
includes the exact rule in the verifier-error feedback (e.g., ``$X$
must cite $Y$ evidence''), does not reliably resolve the routing
error. Mechanical Gate~4 enforcement catches this class of LLM
limitation structurally.

\begin{table}[h]
\centering
\footnotesize
\setlength{\tabcolsep}{2pt}
\begin{tabularx}{\columnwidth}{@{}Xrrr@{}}
\toprule
\textbf{Claim type} & \textbf{mini} & \textbf{Llama} & \textbf{Sonnet} \\
\midrule
\texttt{COMPARISON}         & 371 (55.9\%) & 394 (48.3\%) & 324 (33.9\%) \\
\texttt{FEE\_}\linebreak\texttt{JUSTIFICATION} & 185 (27.9\%) & 154 (18.9\%) & 268 (28.1\%) \\
\texttt{CAP\_SWITCH}        &  21 ~(3.2\%) & 136 (16.7\%) & 220 (23.0\%) \\
\texttt{ASSUMPTION}         &  87 (13.1\%) & 131 (16.1\%) & 143 (15.0\%) \\
\midrule
\textbf{Total error msgs}   & 664           & 815           & 955           \\
\bottomrule
\end{tabularx}
\caption{\textbf{Gate~4 verifier error messages} by claim type
(grounded variant, $r{=}1$, 150 instances per model). The cell counts
report verifier error \emph{messages}, not distinct failing claims --
this is the key distinction from Table~\ref{tab:main}, whose Gate~4
column reports one count per distinct failing claim and is authoritative
for the headline totals. \texttt{CAP\_SWITCH} can generate up to three
messages per claim because \texttt{CAP\_HIT} is checked on two verifier
paths and \texttt{ALLOCATION\_SEGMENT} on one. This is why the totals exceed the
Table~\ref{tab:main} Gate~4 counts (e.g., Sonnet: 955 messages from
841 failing claims). \texttt{COMPARISON} (missing
\texttt{WINNER\_BY\_CATEGORY}) is the largest sub-mode on every
model.}
\label{tab:gate4-submode}
\end{table}

\refstepcounter{section}\label{app:prompts}\section*{Appendix~\Alph{section}. Prompt Templates and Sample Evidence Graph}

This appendix gives the verbatim system-prompt skeletons used in our
runs and a redacted example evidence graph snippet for one instance.
The \verityartifact{} contains the full prompt source and reported per-instance
logs; the excerpts here expose the high-signal portions of
the prompt-engineering choices in the paper.

\paragraph{Grounded narrator prompt (skeleton).}
\begin{quote}\small\ttfamily
You must output ONLY a valid JSON array of\\
claim objects. Cite only the evidence IDs\\
listed below.\\[2pt]
Each claim must have these fields:\\
\phantom{XX}- claimId (string)\\
\phantom{XX}- claimType (one of: COMPARISON, THRESHOLD,\\
\phantom{XX}\hspace{0.6em} ALLOCATION, ASSUMPTION,\\
\phantom{XX}\hspace{0.6em} FEE\_JUSTIFICATION, CAP\_SWITCH)\\
\phantom{XX}- text (string)\\
\phantom{XX}- citedEvidenceIds (array of strings)\\
\phantom{XX}- citedEntities (array of card ids or\\
\phantom{XX}\hspace{0.6em} category names)\\
\phantom{XX}- citedNumbers (array of number strings)\\
Output ONLY a JSON array.\ No commentary.\\[2pt]
Evidence IDs you may cite (exact strings):\\
\phantom{XX}\textit{\{evidenceIdsLine\}}\\[2pt]
For citedEntities, use ONLY these exact\\
strings (card ids, category names, currencies):\\
\phantom{XX}\textit{\{allowedEntitiesLine\}}\\[2pt]
For citedNumbers, use ONLY these exact strings\\
(do not round, do not reformat; small ints\\
0--3 are also allowed):\\
\phantom{XX}\textit{\{allowedNumbersLine\}}\\[2pt]
Structured evidence (for reference):\\
\phantom{XX}\textit{\{evidenceJson\}}
\end{quote}

\paragraph{Ungrounded narrator prompt (skeleton).}
The ungrounded variant differs only in what evidence is supplied:
\begin{quote}\small\ttfamily
Explain why this credit card portfolio is\\
a good choice. Be specific. Cite card names,\\
category earn rates, dollar amounts, and\\
supporting reasons.\\[2pt]
\textit{\{same CLAIM\_SCHEMA block as grounded\}}\\[2pt]
Portfolio summary:\\
\phantom{XX}\textit{\{goal, portfolio card ids,}\\
\phantom{XX}\textit{ primaryCurrency, netValueUsd,}\\
\phantom{XX}\textit{ totalEarnValueUsd, totalCreditValueUsd,}\\
\phantom{XX}\textit{ totalFeesUsd\}}
\end{quote}

The grounded variant gives the LLM the full evidence graph plus
verbatim allowed-entity and allowed-number lists; the ungrounded
variant gives only top-line breakdown numbers (net, earn, credits,
fees) and the portfolio card list. Neither prompt instructs the LLM
to ``hallucinate'' or ``be careful''; the difference is entirely the
evidence provided.

\paragraph{Repair prompt (skeleton).}
On a repair pass, the original grounded prompt is reused with up to
ten specific verifier errors appended:
\begin{quote}\small\ttfamily
\textit{\{original grounded prompt\}}\\[2pt]
Your previous output failed the claim verifier\\
with these errors:\\
\phantom{XX}- citedEntity not allowed: amex-gold\\
\phantom{XX}- citedNumber not in allowed set: 4\\
\phantom{XX}- claim type COMPARISON requires citation\\
\phantom{XX}\hspace{0.6em} of WINNER\_BY\_CATEGORY evidence\\
\phantom{XX}\textit{(up to 10 errors shown)}\\[2pt]
Produce a corrected JSON array. Cite only\\
evidence ids, entities, and numbers listed\\
in the prompt above. Use exactly the strings\\
provided.
\end{quote}

\paragraph{Judge prompt (skeleton, used in \S\ref{sec:judge}).}
\begin{quote}\small\ttfamily
You are an impartial judge comparing two\\
explanations of a credit card portfolio\\
recommendation. Choose the explanation that\\
is most useful to a user reading it.\\
Consider:\\
\phantom{XX}- Faithfulness to the context\\
\phantom{XX}- Clarity\\
\phantom{XX}- Helpfulness\\
Do not let length or position influence you.\\[2pt]
Output ONLY a single JSON object on one line.\\
Schema: \{"winner": "A" | "B" | "TIE",\\
\phantom{XX}"rationale": "<one short sentence>",\\
\phantom{XX}"confidence": <float 0.0 to 1.0>\}\\[2pt]
Context (the optimizer's actual decision):\\
\phantom{XX}\textit{\{contextSummary: goal + portfolio}\\
\phantom{XX}\textit{ ids only; no evidence graph\}}\\[2pt]
Explanation A: \textit{\{prose A\}}\\
Explanation B: \textit{\{prose B\}}
\end{quote}

Note that the judge sees \emph{only the prose} of the two explanations
plus the (goal, portfolio) header. It has no access to the evidence
graph, the verifier output, or which explanation is grounded; this is
why we report the judge result as a usefulness-preference diagnostic
rather than a faithfulness measurement.

\paragraph{Sample evidence graph excerpt.}
A representative grounded instance emits an evidence graph with the
following shape (entity ids replaced with public-domain placeholders
for the excerpt):

\begin{quote}\small\ttfamily
\{ \\
\phantom{XX}graphDigest: ``sha256:5af1\ldots'', \\
\phantom{XX}nodes: [ \\
\phantom{XXXX}\{id:``ev\_1'', type:``ALLOCATION\_SEGMENT'', \\
\phantom{XXXXXX}\hspace{0.5em} card: \\
\phantom{XXXXXXXX}\hspace{0.5em} ``chase-freedom-flex'', \\
\phantom{XXXXXX}\hspace{0.5em} category:``DINING'', spend:``1200.00''\}, \\
\phantom{XXXX}\{id:``ev\_2'', type:``WINNER\_BY\_CATEGORY'', \\
\phantom{XXXXXX}\hspace{0.5em} category:``DINING'', \\
\phantom{XXXXXX}\hspace{0.5em} winner: \\
\phantom{XXXXXXXX}\hspace{0.5em} ``chase-freedom-flex'', \\
\phantom{XXXXXX}\hspace{0.5em} runnerUp:``amex-gold''\}, \\
\phantom{XXXX}\{id:``ev\_3'', type:``FEE\_BREAK\_EVEN'', \\
\phantom{XXXXXX}\hspace{0.5em} card: \\
\phantom{XXXXXXXX}\hspace{0.5em} ``chase-sapphire-preferred'', \\
\phantom{XXXXXX}\hspace{0.5em} fee:``95.00'', breakEven:``1267.00''\}, \\
\phantom{XXXX}\textit{\ldots\ {(typical instance has 14--22 nodes)}} \\
\phantom{XX}], \\
\phantom{XX}allowedEntities: [chase-freedom-flex, \\
\phantom{XXXX}\hspace{0.5em} chase-sapphire-preferred, DINING, \\
\phantom{XXXX}\hspace{0.5em} GROCERIES, TRAVEL, \ldots\ (30 strings)], \\
\phantom{XX}allowedNumbers: [0, 1, 2, 3, 95.00, 197.64, \\
\phantom{XXXX}\hspace{0.5em} 1031.00, 1200.00, 1267.00, \ldots\\
\phantom{XXXX}\hspace{0.5em} (typically 80--110 strings)] \\
\}
\end{quote}

The graph digest is computed by canonicalizing the node list (sorted
by id, fields in fixed order, all numbers as decimal strings) and
hashing with SHA-256. The verifier's accept/reject decision is a
deterministic function of (claim, graph) and is reproducible across
runs with bit-identical input.

\refstepcounter{section}\label{app:prereg}\section*{Appendix~\Alph{section}. Pre-Registration Summary}

The released \path{protocol/eval_preregistration.md} separates these pre-run
decisions from later analysis and deployment choices:

\begin{itemize}\setlength{\itemsep}{2pt}
  \item \textbf{Co-primary metrics:}
    (a) \emph{claim survival} = passing / totalClaims per cell, where the
    implemented totalClaims counts parsed-and-retained claims; (b) mean
    shipped verified claims per instance under permissive emission.
  \item \textbf{Pairing units.}
    Repair-effect contrasts: paired by (profileId, goal, modelId).
    Cross-model contrasts at $r{=}1$: paired by (profileId, goal).
    Sonnet $r{=}2$ ablation: paired by (profileId, goal).
  \item \textbf{Statistical procedure.}
    Instance-cluster bootstrap (10{,}000 iterations, seed~42); report
    paired differences with 95\% CIs (significant iff excluding zero).
  \item \textbf{Run and decision rules.}
    Compare $r{=}0$ with $r{=}1$; claim a repair benefit only if a paired
    survival CI excludes zero. The pre-registered diminishing-returns rule
    used the Sonnet smoke $r{=}2-r{=}1$ CI, which crossed zero at $N{=}9$.
  \item \textbf{Stop conditions.}
    Abort a run above 5\% provider-call failures or 50\% above its cost
    budget; diagnose before rerunning the affected cell.
  \item \textbf{Explicit non-claims} committed up-front:
    \begin{itemize}\setlength{\itemsep}{0pt}
      \item No claim of state-of-the-art on any open-domain
        factuality / RAG benchmark.
      \item No claim of empirical cross-domain transfer or seed robustness.
      \item No unqualified claim that grounding beats the ungrounded variant.
    \end{itemize}
\end{itemize}

\paragraph{Post-registration analysis and deployment choices.}
The per-instance best-of-($r{=}0,r{=}1$) deployment policy, the first-150
lexicographic judge sample, and the large-run Sonnet $r{=}2$ ablation were
specified later. Table~\ref{tab:main} and the main figures instead report
$r{=}0$ and $r{=}1$ separately. The observed 1.78\% schema-failed instance
rate at $r{=}1$ is descriptive; the 5\% stop rule concerned provider calls.
The judge is a usefulness diagnostic, and permissive emission is a filtering
policy, not evidence that the generator improves.

\refstepcounter{section}\label{app:lineage}\section*{Appendix~\Alph{section}. Structured-Data and Citation-Metric Lineage}

Slot-error and semantic-control work checks missing or spurious
meaning-representation values
\cite{wen-etal-2015-semantically,dusek-jurcicek-2016-sequence,juraska-etal-2018-deep,nie-etal-2019-simple,dusek-etal-2019-semantic};
IE and PARENT evaluate table-to-text content
\cite{wiseman-etal-2017-challenges,dhingra-etal-2019-handling}, while
ALCE separates citation correctness and completeness
\cite{gao-etal-2023-enabling}. \textsc{VerityGate} adds a typed
declaration/evidence contract plus repair-and-filter evaluation, not a
new set-membership operation. Neural-metric and relaxed-rule comparisons
remain future work.

\end{document}